\documentclass[iicol,sn-mathphys-num]{sn-jnl}

\usepackage{graphicx}%
\usepackage{multirow}%
\usepackage{amsmath,amssymb,amsfonts}%
\usepackage{amsthm}%
\usepackage{mathrsfs}%
\usepackage[title]{appendix}%
\usepackage[table]{xcolor}
\usepackage{float} 
\usepackage{textcomp}%
\usepackage{manyfoot}%
\usepackage{booktabs}%
\usepackage{cleveref}
\usepackage{ragged2e}
\usepackage{algorithm}%
\usepackage{algorithmicx}%
\usepackage{algpseudocode}%
\usepackage{listings}%
\usepackage{diagbox}%
\usepackage{tabularx}%
\usepackage{array}%
\usepackage{booktabs}%
\usepackage{adjustbox}%
\usepackage{microtype}%
\usepackage{makecell}%
\usepackage{hyperref}
\usepackage{xurl}
\usepackage{lmodern}
\usepackage{graphicx}
\usepackage{flushend}
\usepackage[font=footnotesize]{caption}
\usepackage[font=footnotesize]{subcaption}

\usepackage{rotating} 
\usepackage{textgreek}
\usepackage{tikz}
\usetikzlibrary{positioning}
\usepackage{tabularx}
\usepackage{array}
\usepackage{threeparttable}
\usepackage{makecell}
\usepackage{longtable}
\usepackage{adjustbox}
\usetikzlibrary{shapes.geometric, arrows.meta, positioning}
\newcolumntype{L}[1]{>{\RaggedRight\arraybackslash}p{#1}}
\usetikzlibrary{arrows.meta, positioning}
\tikzset{main/.style={circle, draw, fill=blue!20, thick, minimum size=2cm, text centered}}
\Crefname{figure}{Fig.}{Figs.} 

\renewcommand{\figurename}{Fig.}
\usepackage[labelfont=bf]{caption}
\DeclareCaptionLabelSeparator{twolines}{\newline}
\usepackage{setspace}
\usepackage{siunitx} 
\newcolumntype{M}{>{\centering\arraybackslash}m{0.25\columnwidth}}

\usepackage{csquotes}
\usepackage{ragged2e}

\usepackage{stfloats} 

\begin{document}

\title[Article Title]{
Human-Centered Reflections on Care Robots: A Comparative Study of Caregiver Perspectives}


\author[1,2]{{Laura} {Londoño}}
\email{londono@cs.uni-freiburg.de}

\author[3]{{Klaus} {Baumann}}

\author[2]{{Abhinav} {Valada}}

\author[1]{{Markus} {Langer}}

\affil[1]{\orgdiv{Department of Psychology}, \orgname{University of Freiburg}, \country{Germany}}
\affil[2]{\orgdiv{Department of Computer Science}, \orgname{University of Freiburg}, \country{Germany}}
\affil[3]{\orgdiv{Department of Theology}, \orgname{University of Freiburg}, \country{Germany}}


\abstract{Care robots are increasingly being introduced into healthcare settings, raising important questions about their acceptance and ethical implementation. To better understand these challenges, this study investigates caregivers’ perceptions of four categories of care robots: delivering supplies, helping patients into bed, monitoring vital signs, and assisting with mobility. We conducted a mixed-methods study employing a mixed-factorial design in which 298 caregivers from the United States, Mexico, and Chile evaluated all four robot categories. Quantitative measures integrated constructs from the Unified Theory of Acceptance and Use of Technology, the Cognitive–Affective–Normative model, and overall acceptance ratings. Qualitative data were collected through open-ended questions and analyzed using a literature-informed ethical framework. The results indicate that participants across countries generally evaluated care robots positively, particularly for logistical and physically demanding tasks rather than those requiring intensive interpersonal interaction. The qualitative findings provide further insight into stakeholders' views of the ethical implications of care robot use. Participants emphasized potential benefits such as reduced workload, lower risk, and greater patient autonomy, while also expressing concerns about dependability, the need for human oversight, and potential job displacement. Although many ethical concerns were shared across countries, participants differed in how they interpreted and prioritized them. These findings advance a context-sensitive and socially informed understanding of responsible design and implementation of care robots.}




\maketitle

\section{Introduction}\label{sec1}


The use of robots has expanded beyond industrial applications into a range of diverse social domains, including healthcare and long-term care~\citep{silvera2024robotics,Sequeira2018,mohan2024syn,hindel2025dynamic}. As robots become increasingly integrated into those settings characterized by vulnerability and dependence, their presence raises critical questions about how they should interact with caregivers, support patients, and participate in practices traditionally grounded in human relationships~\citep{de2013exploring, johansson2022significant,frennert2020expectations,johansson2020care}. Addressing these questions requires more than improving robots’ technical performance and usability. It also demands careful consideration of the ethical, social, cultural, and emotional dimensions of care.   


As part of broader efforts to promote the responsible development of care robots and foster their acceptance, scholars have emphasized the importance of incorporating the perspectives of stakeholders directly involved in the design, deployment, and use~ \citep{bruno2019knowledge,sanoubari2018explicit, broadbent2009acceptance,londono2022doing,livanec2025designing}. For example, \citep{johansson2020care, Klebbe2025RoleOfRobots} argues that meaningful acceptance requires a thorough understanding of why end users accept or reject care robots in social contexts. They argue that responsible and context-sensitive care robotics should align robots' technical capabilities and forms of agency with the goals, values, and needs of professional care practices.

Despite growing interest in the ethical design and acceptance of care robots, stakeholders from countries outside the so-called “developed world” remain underrepresented in the literature~\citep{bon2022decolonizing, garcia2025technology, gordon2022development}. This imbalance may introduce systematic biases into the development of care technologies by privileging culturally narrow assumptions about care practices, technological expectations, and user needs. Moreover, acceptance and ethical concerns may depend not only on sociocultural context but also on the specific function assigned to a robot. A robot that transports supplies, for example, may be evaluated according to different expectations and ethical criteria than one that physically assists or monitors a patient. 

To address this gap, this study investigates (1)~caregivers’ perceptions of four categories of care robots: robots that deliver supplies, help patients into bed, monitor vital signs, and assist with mobility. The study compares stakeholders from the United States, Mexico, and Chile, three countries with distinct socioeconomic, cultural, institutional, and technological contexts; and (2) it examines both the acceptance of these robot categories and the ethical considerations that stakeholders regard as important for their responsible design and deployment. 

Our analysis is guided by three premises: (1) acceptance cannot be understood independently of ethical considerations, as perceptions of a robot's benefits, risks, and impact on care shape users' willingness to adopt it; (2) a cross-country perspective is needed to understand how social, cultural, and institutional contexts influence these perceptions; and (3) acceptance is task-specific, meaning evaluations depend on the robot's function and care context and cannot be readily generalized across applications.\looseness=-1

\begin{figure}
    \centering
    \includegraphics[width=\columnwidth]{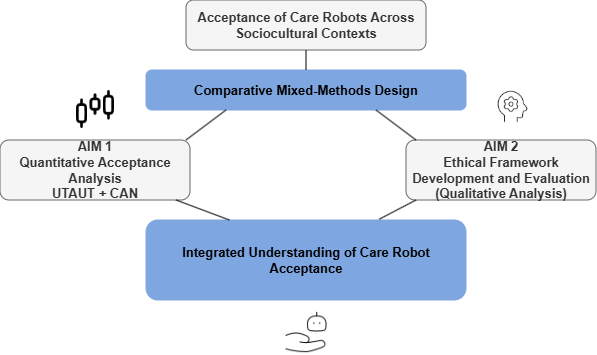}
    \caption{Overview of the comparative mixed-methods design. Quantitative acceptance analyses (UTAUT and CAN) and qualitative ethical evaluations were conducted in parallel and subsequently integrated to develop a comprehensive understanding of the acceptance of four types of care robots across sociocultural contexts.}
    \label{fig:aims-flowchart}
\end{figure}

 
Accordingly, this study pursues two main objectives (see \autoref{fig:aims-flowchart}):
\begin{itemize}
    \item \textbf{Aim 1:} To experimentally evaluate caregivers’ acceptance of four categories of healthcare robots across three countries (the United States, Mexico, and Chile) with distinct socioeconomic and technological contexts. Acceptance is assessed using constructs derived from the Unified Theory of Acceptance and Use of Technology (UTAUT)~\citep{venkatesh2003user} and Cognitive–Affective–Normative (CAN)~\citep{ReinaresLara2018} model, together with an overall acceptance rating. The analysis identifies factors associated with acceptance and compares acceptance profiles across robot categories and countries.  
    \item \textbf{Aim 2:} To develop and apply a literature-informed ethical framework for examining the considerations that caregivers regard as important for the responsible design and deployment of the four categories of care robots across different sociocultural contexts.
\end{itemize}

To address the objectives, we employ a mixed-methods study that combined quantitative and qualitative approaches within a mixed-factorial framework. Robot category served as the within-subjects factor, as each participant evaluated all four categories of care robots. Sociocultural context served as the between-subjects factor and was operationalized through participants’ country of residence and language. 
Participants completed an online survey in which the four robot categories, each representing distinct clinical functions, were presented in randomized order. The quantitative component collected standardized measures of acceptance for each category. The qualitative component used open-ended questions to elicit participants’ views on the ethical considerations relevant to the design and deployment of care robots. These responses  enabled us to identify recurring ethical themes that standardized quantitative measures might not capture. Integrating the quantitative and qualitative findings allowed us to examine not only whether stakeholders accepted different types of care robots, but also how they understood the ethical considerations associated with care robots. 

Accordingly, this study makes three contributions. First, it provides a within-participant comparison of caregivers’ evaluations of four care-robot scenarios representing distinct clinical functions. Second, it integrates quantitative acceptance measures with qualitative ethical analysis to identify how concerns relating to safety, autonomy, relational care, accountability, and accessibility shape and qualify participants’ evaluations of these technologies. Third, it offers comparative evidence on similarities and differences between the sampled groups, comprising English-language participants in the United States and Spanish-language participants in Mexico and Chile.

\section{Related Work}

This section provides a focused review of the literature on technology acceptance. It also examines key findings on ethical evaluations of robot design, as well as research highlighting the importance of robot type and task characteristics in shaping user acceptance. In addition, the section considers cross-cultural variations in the adoption and acceptance of robotics technology. Finally, it analyzes the healthcare contexts of the United States, Mexico, and Chile to situate the study’s findings within their respective sociocultural and structural conditions.

\subsection{Technology Acceptance Models}

Technology acceptance has been extensively studied across multiple disciplines, leading to the development of numerous theoretical frameworks and standardized measurement instruments. Among the most influential is the Unified Theory of Acceptance and Use of Technology (UTAUT) \citep{venkatesh2003user}, which integrates and extends the Technology Acceptance Model (TAM) \citep{holden2010technology} and other earlier frameworks. UTAUT provides a comprehensive account of technology adoption by incorporating social, organizational, and contextual factors that influence both behavioral intention and actual technology use. It proposes four core constructs: performance expectancy, effort expectancy, social influence, and facilitating conditions. This model recognizes that their influence varies according to demographic and situational factors such as age, gender, prior experience, and voluntariness of use \citep{roy2025applying}. 

In addition to technology acceptance models, other frameworks can contribute to the evaluation of emerging technologies by capturing dimensions beyond instrumental acceptance. One example is the Cognitive–Affective–Normative (CAN) model proposed by \citet{ReinaresLara2018}, which incorporates ethical considerations by assessing users' perceptions of fairness, morality, and other normative aspects of technology evaluation. While the CAN model enables the measurement of ethical perceptions, it offers limited insight into the underlying values and forms of ethical reasoning that shape these judgments. This limitation motivates the use of complementary qualitative approaches to better understand stakeholders' ethical evaluations of emerging technologies.

\subsection{Ethical Evaluation of Care Robots}

While ethical reflection surrounding care robots has received increasing attention within the fields of responsible robotics, much of the existing literature has focused on developing normative frameworks, ethical principles, and design guidelines \citep{torras2024ethics,yuan2023ethical,leineweber2025ethical,londono2024fairness}. Comparatively little attention has been devoted to understanding how these normative insights can be integrated with empirical investigations or how ethical considerations shape the acceptance, rejection, and evaluation of care robots in practice \citep{vandemeulebroucke2018use,van2020designing, van2020designing, Hung2022TechnologicalRisks}. As a result, a persistent divide remains between normative ethical theorizing and empirical research \citep{vandemeulebroucke2020ethics}. 

In response, there has been growing recognition of the need for research approaches that move beyond this traditional separation \citep{leineweber2025ethical, Klebbe2025RoleOfRobots, roy2025applying}. Among the most influential contributions in this area is the work of \citet{van2020designing}, who develops an ethical framework for evaluating the design, implementation, and use of care robots. Starting from the question of how robots can support care without undermining the ethical values that make care meaningful, the author argues that care should not be reduced to the automation of care-related tasks. Drawing on the ethics of care tradition developed by \citet{gilligan1993different} and \citet{tronto2015whocares}, she argues that care is fundamentally a relational moral practice rather than a specific activity or set of tasks. Consequently, the ethical evaluation of care robots must consider not only their effectiveness and precision in performing care functions but also their impact on the relationships, responsibilities, and values that constitute caring practices. 

While \citet{van2020designing} focus primarily on normative guidelines for the ethical design of care robots, \citet{vandemeulebroucke2020ethics} argue that ethical evaluations must also consider the socio-historical, organizational, and institutional contexts in which these technologies are implemented. From this perspective, care robots should be evaluated not only in terms of their technical capabilities but also in relation to the broader care systems and social conditions in which they operate. Building on this work, the present study proposes a theoretical framework that integrates multiple ethical perspectives and applies them to the analysis of empirical data, thereby responding to calls for approaches that bridge normative ethics and empirical research.

\subsubsection{Ethical Foundations of the Analytical Framework}
\label{sec:ethical-perspectives}

The theoretical framework proposed by \citet{van2020designing} provides an important foundation for understanding the relational dimensions of care through the lens of care ethics. However, in developing the ethical framework guiding the analysis in the present study, we considered that relying solely on this perspective would be insufficient to capture the broader range of ethical concerns that participants might articulate. To address this limitation, we incorporated additional ethical theories to enable a more comprehensive analysis of the diverse values that shape perceptions of care robots. 

Following, we present a short description of the ethical theories that guide our framework: 

\begin{itemize}
    \item \hypertarget{care-ethics}{\textbf{Care Ethics}}:
    Originally developed by \citet{gilligan1993different} as a critique of justice-based moral theories, the ethics of care emphasizes the moral significance of relationships, interdependence, and contextual responsibilities. From this perspective, ethical judgment arises from attentiveness to concrete relational contexts and the needs of others, rather than from the application of abstract universal rules. Core values associated with this perspective include care, responsibility, relationality, empathy, interdependence, and sensitivity to context.

    \item \hypertarget{capability-approach}{\textbf{Capability Approach}}:
    Developed by \citet{nussbaum2000women}, this framework emphasizes enabling individuals to achieve fundamental capabilities, such as health, bodily integrity, autonomy, emotional development, and control over their environment and decisions. Ensuring these capabilities is considered an ethical obligation, as they constitute the minimum conditions required for a dignified human life. Key values associated with this perspective include human dignity, agency, freedom of choice, equality of capabilities, attention to diversity, and social justice.

    \item \hypertarget{consequentialism}{\textbf{Consequentialism}}:
    Consequentialist theories evaluate the moral rightness of actions primarily in terms of their outcomes \citep{sinnott2015consequentialism,floridi2022unified}. From this perspective, actions are assessed according to the extent to which they produce desirable consequences or minimize harm. Core values associated with this perspective include the moral primacy of consequences, the maximization of overall good, impartiality, aggregation of outcomes, forward-looking moral reasoning, and sensitivity to cost--benefit considerations.
\end{itemize}

The theoretical framework was developed by drawing on these ethical theories to address three important limitations in the existing literature. First, while technology acceptance models are effective in capturing perceptions of usefulness, attitudes, and adoption intentions, they provide only limited insight into the ethical reasoning that underlies these evaluations. Second, many normative approaches to care robotics rely on a single ethical perspective, thereby overlooking the diversity of values, principles, and concerns that stakeholders may express. Third, there is a need for an ethical framework that can be applied to the analysis of empirical data, enabling a more systematic integration of normative ethical analysis with empirical research.

\subsection{Robot Type and Task Characteristics}

In the context of social robots, acceptance also depends on robot-related characteristics, including embodiment, social capabilities, perceived agency, and emotional expressiveness \citep{walters2008avoiding}. In addition, both the type of robot and the tasks it performs influence how it is perceived and accepted across contexts \citep{chatzoglou2024factors, niemela2019social, Formosa2021RobotAutonomy, Johnston2022EthicalDesign}. Consistent with this, \citet{chatzoglou2024factors} and \citet{degraaf2013exploring} found that robot-related characteristics, such as physical attractiveness and task importance, exert a stronger influence on acceptance than socio-cultural factors. Their findings further suggest that users evaluate robots according to the tasks they perform and the value these tasks provide in everyday life. 

Similar evidence has been reported in the healthcare robotics literature. In their review of healthcare robots for older adults, \citet{broadbent2009acceptance} identified the robot’s intended role and function as important determinants of acceptance. The authors argued that users may evaluate robots differently depending on the tasks they are designed to perform, ranging from health monitoring and reminders to physical assistance and social companionship. Although these conclusions were based on a literature review rather than direct empirical comparisons across robot types, the review provided early evidence that task characteristics influence robot acceptance and highlighted the need for systematic empirical research.

Taken together, this body of research suggests that social robot acceptance is task-dependent. Users evaluate robots according to the functions they perform and the value they provide in a given context. Consequently, findings from one robotic application cannot necessarily be generalized to others, highlighting the need to examine multiple robot types performing different care-related tasks. 

\subsection{Cross-Cultural Differences in Care Robot Adoption and Implementation}

Several scholars argue that, despite the transformative impact of digital technologies, many people, particularly in countries with limited participation in technological development, remain excluded from the debates shaping digital societies, a phenomenon referred to as \textit{digital coloniality} \citep{bon2022decolonizing, Leterme2020NorthSouthDigital}. Because digital societies reflect the social structures in which they are embedded, they often reproduce existing historical inequalities \citep{Yu2023AIDivide}. As digital innovation is concentrated in the Global North, the priorities, values, and assumptions embedded in digital technologies predominantly reflect the perspectives of these regions, reinforcing structural imbalances and marginalizing voices from elsewhere.

One way to address this challenge is to design robots that can adapt to diverse socio-cultural contexts. Research in human–robot interaction has shown that robots developed within specific sociocultural settings often embed implicit assumptions about interpersonal distance, communication styles, and social roles \citep{hurtado2021learning, ahmad2025embodiment, ferrer2013social, ferrer2017robot}. When treated as universal interaction norms, these assumptions may conflict with the social practices of other cultural contexts \citep{lawrence2025socialnorms, ahmad2025embodiment}. Moreover, \citet{lawrence2025socialnorms} argue that social norms in human–robot interaction should be understood not only at the societal level but also at the individual level, where personal expectations shape how robot behavior is interpreted and evaluated. Consistent with this view, \citet{marchesi2021cultural} found that individual cultural values, rather than nationality, better predict the social inclusion of robots. 

In contrast, \citet{li2019perceptions} found that expectations regarding appropriate robot behavior vary across cultural contexts. In their cross-cultural experiment with participants from the United States and China, preferences differed with respect to robot autonomy: while participants in some contexts preferred more autonomous and proactive robots, others expected robots to behave more passively. Together, these findings highlight the importance of examining social robot acceptance across different cultural contexts, as acceptance is a key prerequisite for the successful deployment of robots and their contribution to stakeholder well-being.

\subsection{Healthcare and Sociocultural Contexts in the United States, Mexico, and Chile}

Healthcare environments across nations are shaped by a range of structural factors that influence how innovations are developed, adopted, and integrated \citep{kovner2011jonas}. Given that the acceptance and implementation of healthcare robots are likely to be influenced by these contextual conditions, this study focuses on three distinct national settings—the United States, Mexico, and Chile. The following outlines key characteristics of these healthcare environments, with particular attention to technology adoption and implementation.

\subsubsection{Technology Adoption in the United States Healthcare System}

In recent decades, healthcare systems in the United States have been characterized by a strong orientation toward technological innovation \citep{matheny2020artificial}. Empirical evidence indicates sustained growth in the deployment of surgical robots and artificial intelligence–based systems across hospitals, supported by regulatory approvals and institutional adoption patterns that facilitate their incorporation \citep{lee2024autonomy}. By 2022, nearly one-fifth of U.S. hospitals had adopted some form of artificial intelligence \citep{abdullah2024aihospital, cruz2024hospitalrobotics}.\looseness=-1

As a result, contemporary hospital environments increasingly involve the coexistence of caregivers, patients, and automated or semi-automated technological systems. This growing technological density not only reshapes clinical practices and organizational processes but also contributes to evolving expectations regarding efficiency, precision, and the role of technology in medical decision-making. However, despite the rapid implementation of advanced technologies in healthcare settings in the U.S., the adoption of AI remains uneven across institutions \citep{abdullah2024aihospital}. 

The implementation of robotic systems in healthcare and care settings has also reshaped how caregivers conceptualize work organization, care relationships, and patient safety \citep{soljacic2024robots}. While these technologies are often promoted as tools for improving efficiency, reducing workload, and optimizing care processes, their integration has also generated significant ethical and professional concerns. In particular, scholars have questioned whether an increasing reliance on technological solutions may contribute to the reduction of care to a set of technical tasks, potentially undermining the relational and emotional dimensions that are central to caregiving \citep{soljacic2024robots}. 

\subsubsection{Technology Adoption in Mexican and Chilean Healthcare Systems}

Mexico and Chile were grouped to represent a broader Latin American context characterized by comparatively lower levels of healthcare robot implementation, greater resource constraints, and sociocultural values surrounding care that may differ from those commonly described in the US. These countries also remain largely dependent on technological innovations developed in countries like the US \citep{schneegans2021race,dutta2024understanding}. This situation is primarily driven by comparatively low levels of investment in technological development, limited scientific infrastructure, and a continued reliance on imported technologies \citep{gereffi2018global}. Such asymmetries in technological production reinforce existing patterns of dependency.

Within this structural context, inequality remains a defining feature of the region, with wealth and social resources concentrated among a small segment of the population \citep{dachs2002inequalities}. These disparities contribute to stratified healthcare systems in which access to infrastructure, specialized personnel, and advanced medical technologies is unevenly distributed across social classes and geographic regions, systematically disadvantaging rural and marginalized communities. Moreover, institutional arrangements, financing mechanisms, and patterns of technological diffusion often reinforce these inequalities by favoring already well-resourced sectors \citep{dachs2002inequalities}. Consequently, technological innovation can have ambivalent effects: while new medical technologies may improve healthcare access, efficiency, and reduce costs, they may also reproduce or exacerbate existing inequalities when not adapted to local socio-economic and institutional conditions \citep{dutta2024understanding}.

Research examining caregivers' perceptions of social robots in Latin America remains limited. One notable contribution in the region is the study by \citet{portugalchurata2026socialrobotacceptance}, which investigated factors shaping the acceptance of social robots among Brazilian caregivers involved in older-adult care. The findings indicate that caregivers generally perceived social robots as valuable tools for supporting instrumental care tasks, particularly those related to safety monitoring and medication management. At the same time, participants expressed concerns regarding technical reliability, maintenance requirements, and data security. Interestingly, acceptance was more strongly associated with caregivers' perceived workload than with prior caregiving experience or familiarity with social robots.

\subsubsection{Research Gaps and Questions}

The related literature provides the theoretical and empirical foundations for addressing the research questions that guide this study. In particular, it reveals two interrelated gaps. First, technology acceptance research has largely emphasized instrumental and functional determinants of adoption, while paying comparatively limited attention to how stakeholders interpret and evaluate the ethical implications of robotic design and implementation. Second, despite the growing body of research on care robotics, comparative studies examining caregivers’ perceptions of care robots across diverse sociocultural contexts remain limited. This is especially evident in comparisons involving stakeholders from highly technologized healthcare environments, such as the United States, and stakeholders from Latin American contexts, such as Mexico and Chile, which differ in terms of technological development. 

\begin{itemize}
\item \textbf{RQ1:} To what extent do acceptance-related perceptions of four categories of healthcare robots, measured using UTAUT, CAN constructs, and an overall rating, vary across three countries with distinct socioeconomic and
technological contexts (the United States, Mexico, and
Chile)?
\item \textbf{RQ2:} What similarities and differences emerge in stakeholders’ understandings of the ethical design and deployment of care robots when analyzed through the proposed literature-based ethical framework, and how can these variations be interpreted in relation to ethical reasoning and values-based considerations?
\end{itemize}

\section{Methods}

This section describes the study design, data collection procedures, and analytical approach.

\subsection{Participants}

A total of 298 participants from the United States (\textit{N} = 152), Mexico, and Chile (\textit{N = 146}), took part in this study. Participants were recruited through the Prolific platform using a purposive sampling strategy. Specific inclusion filters were applied to target individuals residing in the USA, Mexico, or Chile, who were fluent in English (in the case of the United States) or Spanish (in the case of Mexico and Chile), and who reported professional experience or affiliation within the healthcare sector. 

This sampling approach enabled the deliberate selection of participants relevant to the study’s objectives, ensuring that the sample reflected the sociocultural and professional characteristics necessary to examine variations in technology acceptance and ethical reasoning across socio-cultural contexts. The study lasted approximately 27 minutes, and participants were compensated with £10.23.

\subsection{Experimental Setup}

To address the research questions, we employed a mixed-methods design that integrated both quantitative and qualitative analytical approaches within a mixed factorial structure, consisting of one within-subjects factor and one between-subjects factor. The within-subjects factor was robot type (Robots Delivering Supplies, Robots Helping Patients into Bed, Robots Monitoring Vital Signs, and Robots Assisting with Mobility), as all participants saw videos of each robot type and then evaluated each through questionnaires. The between-subjects factor was sociocultural context, operationalized through participants’ country of residence and language, and subsequently grouped into two broader groups: English group (EN) and Mexico/Chile group (SP).  

\begin{figure*}
\centering

\begin{subfigure}[t]{0.24\textwidth}
    \centering
    \includegraphics[width=\linewidth,height=2.7cm]{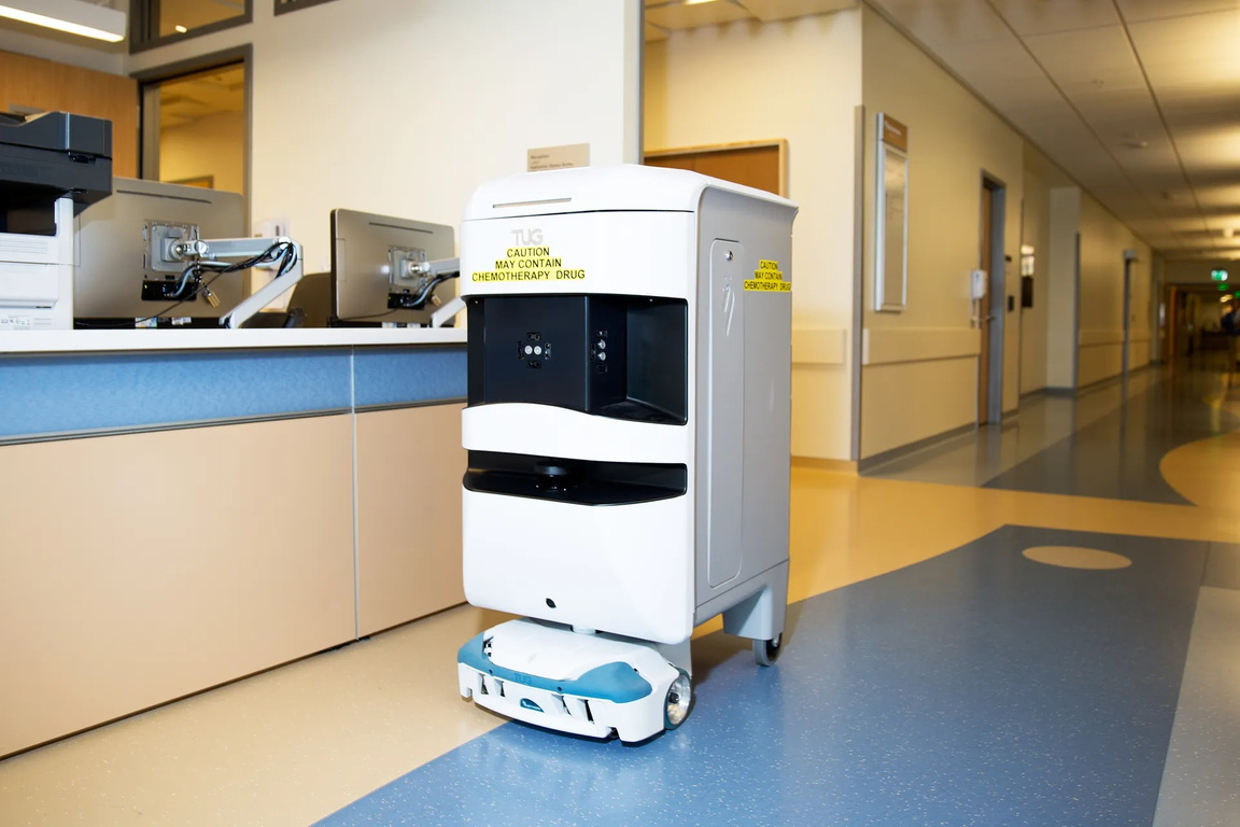}
    \caption{Robot Delivering Supplies}
\end{subfigure}
\hfill
\begin{subfigure}[t]{0.24\textwidth}
    \centering
    \includegraphics[width=\linewidth,height=2.7cm]{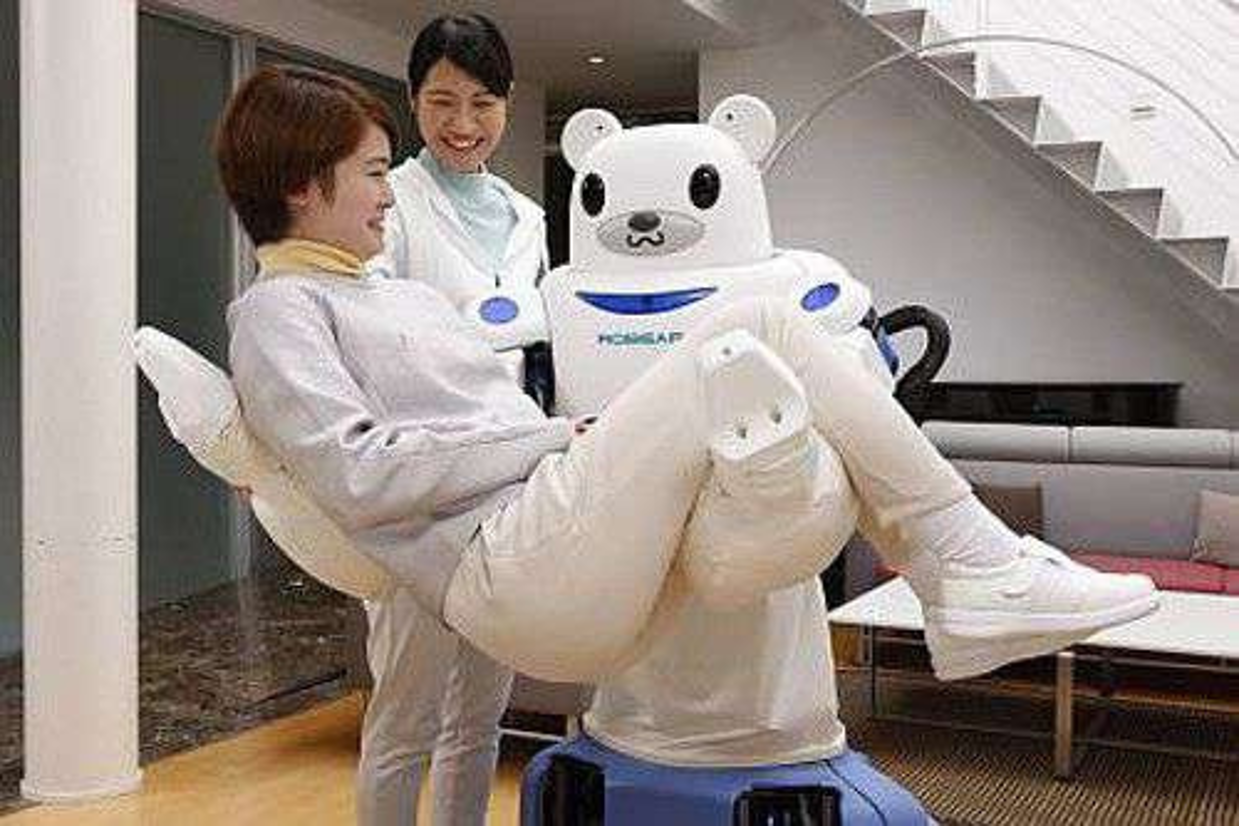}
    \caption{Robot Helping Patients into Bed}
\end{subfigure}
\hfill
\begin{subfigure}[t]{0.24\textwidth}
    \centering
    \includegraphics[width=\linewidth,height=2.7cm]{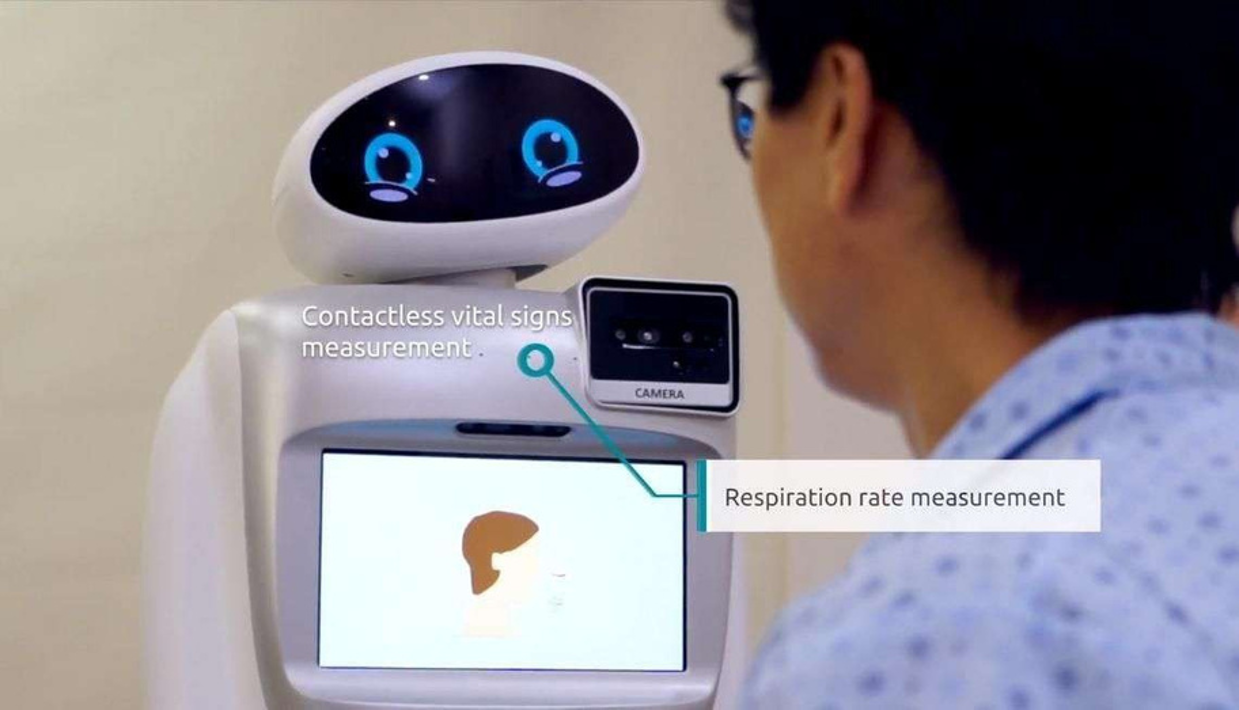}
    \caption{Robot Monitoring Vital Signs}
\end{subfigure}
\hfill
\begin{subfigure}[t]{0.24\textwidth}
    \centering
    \includegraphics[width=\linewidth,height=2.7cm]{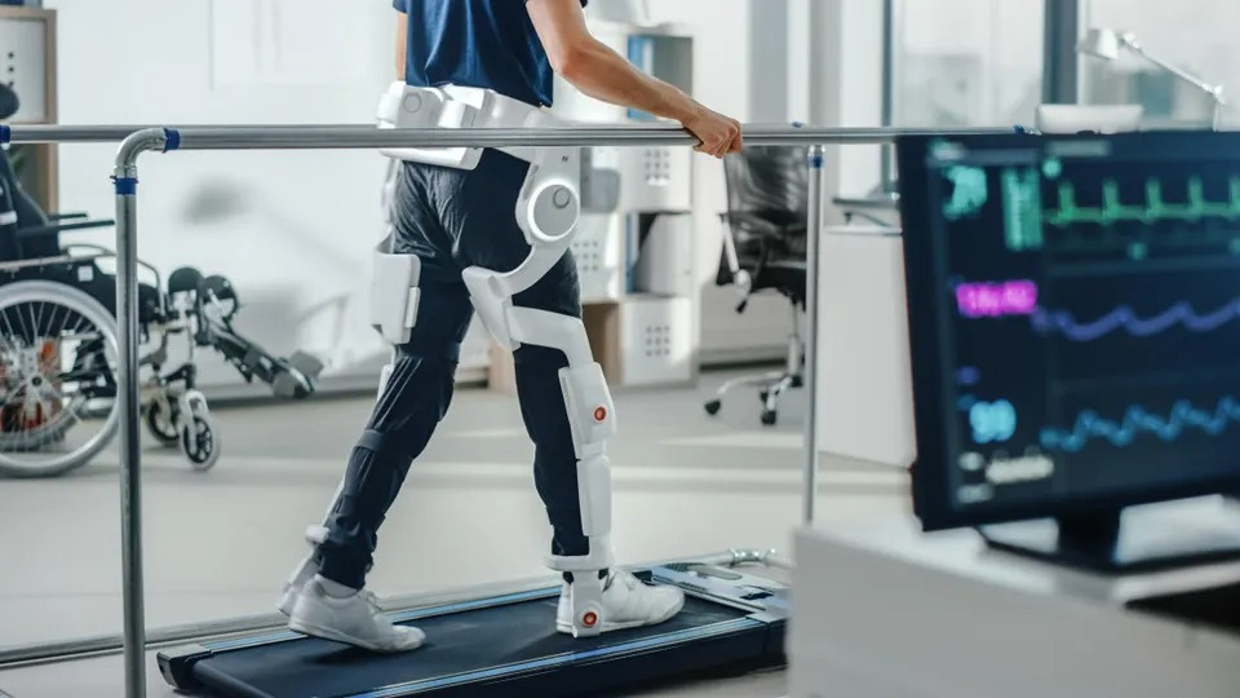}
    \caption{Robot Assisting with Mobility}
\end{subfigure}

\caption{
Representative examples of the four healthcare robot categories included in the study:
(a) autonomous delivery robot (TUG) \cite{wired2015hospitalrobot},
(b) patient-assistive robot (Robear) designed to support patient transfer and lifting tasks \cite{nursingtimes2015robot},
(c) monitoring robot (Florence) used for patient assessment and vital-sign monitoring \cite{govinsider2023robot}, and
(d) rehabilitation exoskeleton designed to support mobility and physical recovery \cite{forbes2023exoskeleton}.
}
\label{fig:healthcare_robot_types}
\end{figure*}

At the beginning of the study, all participants received a general introduction to the concept of care robots. They then received a standardized written description of each of the four categories of the health robots included in the experiment  (see Table~\ref{tab:Typesofrobots}, \autoref{fig:healthcare_robot_types}), followed by a short video clip illustrating the operational functionality of the respective robots. The selected robot types were chosen for two main reasons: 
(1)~they represent distinct and sufficiently differentiated robot functions, allowing for the examination of how variations in robot tasks and decision-making  influence human acceptance; and
(2)~they involve different levels of potential risk and forms of human–robot interaction, thereby enabling an analysis of acceptance across a continuum of ethical and practical concerns.

\begin{figure}[ht]
\centering
\resizebox{0.40\columnwidth}{!}{%
\begin{tikzpicture}[
    node distance=0.45cm,
    startstop/.style={
        rectangle,
        rounded corners,
        minimum width=3.4cm,
        minimum height=0.6cm,
        text centered,
        draw=black,
        fill=gray!15,
        font=\scriptsize
    },
    process/.style={
        rectangle,
        minimum width=3.4cm,
        minimum height=0.6cm,
        text centered,
        draw=black,
        fill=blue!8,
        font=\scriptsize
    },
    robotblock/.style={
        rectangle,
        rounded corners,
        minimum width=3.8cm,
        minimum height=1.0cm,
        text width=3.6cm,
        align=center,
        draw=black,
        fill=green!10,
        font=\scriptsize
    },
    arrow/.style={thick, -{Latex}}
]

\node (welcome) [startstop] {Welcome Study};
\node (consent) [process, below=of welcome] {Informed Consent};
\node (scenario) [process, below=of consent] {Initial Scenario Test};

\node (robot1) [robotblock, below=of scenario] {
\textbf{Robot 1}\\
Description + Video\\
UTAUT/CAN/Overall Rating \\
Open-ended Questions
};

\node (robot2) [robotblock, below=of robot1] {
\textbf{Robot 2}\\
Description + Video\\
UTAUT/CAN/Overall Rating \\
Open-ended Questions
};

\node (robot3) [robotblock, below=of robot2] {
\textbf{Robot 3}\\
Description + Video\\
UTAUT/CAN/Overall Rating \\
Open-ended Questions
};

\node (robot4) [robotblock, below=of robot3] {
\textbf{Robot 4}\\
Description + Video\\
UTAUT/CAN/Overall Rating \\
Open-ended Questions
};

\node (end) [startstop, below=of robot4] {End of Study};

\draw [arrow] (welcome) -- (consent);
\draw [arrow] (consent) -- (scenario);
\draw [arrow] (scenario) -- (robot1);
\draw [arrow] (robot1) -- (robot2);
\draw [arrow] (robot2) -- (robot3);
\draw [arrow] (robot3) -- (robot4);
\draw [arrow] (robot4) -- (end);

\end{tikzpicture}%
}
\caption{Schematic representation of the experimental procedure, including participant onboarding, scenario assessment, and repeated evaluations of the four care robots.}
\label{fig:flowchart}
\end{figure}
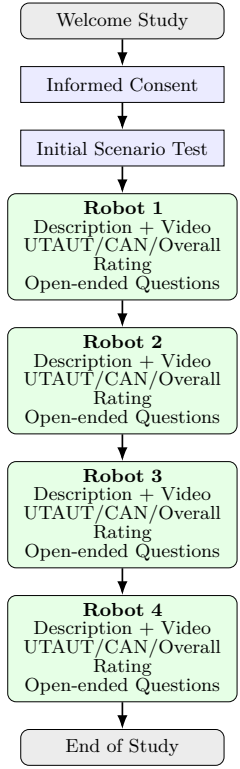

After viewing each video, participants completed a questionnaire consisting of quantitative measures based on the UTAUT and CAN scales, as well as an overall rating. In addition, after each video, participants responded to a set of open-ended questions designed to capture their ethical reflections. \autoref{fig:flowchart} presents a schematic overview of the experimental procedure. Each participant evaluated all videos presented in randomized order, ensuring balanced exposure across conditions and enabling within-subject comparisons. Two parallel versions of the experimental materials were developed: one in English and one in Spanish. Content consistency across both versions was verified by two researchers, one a native Spanish speaker and the other a native English speaker. Each version was accessed through a separate study link.

\begin{table}
\centering
\footnotesize
\caption{Description of the four robot categories used as experimental stimuli.}
\begin{tabularx}{\columnwidth}{|p{2cm}|X|}
\hline
\textbf{Robot Type} & \textbf{Description} \\
\hline

\textbf{Delivering Supplies (DS)} &
Mobile robots that transport medications, sterile equipment, and laboratory samples, reducing nursing workload and improving logistical efficiency, especially under high-pressure conditions or isolation protocols. \\
\hline

\textbf{Helping Patients into Bed (HPB)} &
Robots that assist caregivers in lifting and transferring patients with limited mobility, helping to prevent injuries and physical strain while improving safety and efficiency in patient-handling tasks. \\
\hline

\textbf{Monitoring Vital Signs} &
Robots equipped with sensors to autonomously measure heart rate, blood pressure, respiratory rate, and oxygen saturation, improving accuracy and reducing human error in routine monitoring. \\
\hline

\textbf{Assisting with Mobility} &
Exoskeleton-based or mobility-assistive robots that support rehabilitation by helping patients walk and stand, monitoring posture, balance, and gait, and reducing caregiver workload and risk of falls. \\
\hline
\end{tabularx}
\label{tab:Typesofrobots}
\end{table}

\subsection{Measures}

For the analysis, two primary outcome variables were analyzed: 1) Acceptance of care robots, measured and compared across countries using constructs derived from UTAUT, CAN, and an overall rating; and 2) Ethical and normative considerations examined through a qualitative-theoretical analysis to identify dimensions of ethical reasoning and values-based judgments influencing acceptance. 

\subsection{Survey Instrument and Procedure}

The questionnaire was composed of two sections designed to capture participants' perceptions, evaluations, and reflections. The structure and rationale of each component are described below.

\subsubsection{Quantitative Measures: Selection and Adaptation of Scales}

The questionnaire combined constructs from the UTAUT model \citep{venkatesh2003user} and the CAN model \citep{reinares2018you}. UTAUT captures key determinants of technology acceptance, including Performance Expectancy, Attitudes Toward Using Technology, Self-Efficacy, and Behavioral Intention, whereas the CAN model complements these constructs by incorporating normative dimensions. Given the relational and ethically sensitive nature of care robots, integrating both models enabled a more comprehensive assessment of acceptance. We also included an overall rating.

Because the original UTAUT items were developed for broader technology contexts, minor wording adaptations were introduced to improve ecological validity in healthcare settings. For example, references to "technology" were replaced with "the robot." This adaptation approach is consistent with previous acceptance research \citep{shiferaw2021healthcare,rahimi2018systematic} and preserves the theoretical meaning of the original scales. Table~\ref{tab:scales_tab} summarizes the rationale and items included for each construct.

\newcolumntype{Y}{>{\RaggedRight\arraybackslash}X}

\begin{table*}[t]
\centering
\caption{Overview of the scales used in the user study. For each item, participants indicated their agreement on a 7-point Likert scale.}
\label{tab:scales_tab}

\footnotesize
\renewcommand{\arraystretch}{1.15}
\setlength{\tabcolsep}{5pt}

\begin{adjustbox}{max width=\textwidth}

\rowcolors{2}{gray!4}{white}

\begin{tabularx}{\textwidth}{
>{\raggedright\arraybackslash}p{3.1cm}
>{\raggedright\arraybackslash}p{5.3cm}
>{\raggedright\arraybackslash}X
}

\toprule
\rowcolor{gray!15}
\textbf{Dimension} &
\textbf{Rationale} &
\textbf{Items used in the study} \\
\midrule

\textbf{Performance Expectancy}~\cite{Venkatesh2003}
&
Assesses the perceived usefulness of care robots for supporting clinical work and improving job performance. 
&
1. I would find the robot useful in my job.\\
&
&
2. Using the robot would enable me to accomplish tasks more quickly.\\
&
&
3. Using the robot would increase my productivity.\\
&
&
4. If I used the robot, I would have more chances of getting a raise.\\

\midrule

\textbf{Attitude Toward Using Technology}~\cite{Venkatesh2003}
&
Assesses participants’ affective evaluations of interacting with care robots, including perceived enjoyment, interest, and satisfaction.
&
1. My interaction with the robot would be clear and understandable.\\
&
&
2. The system would make work more interesting.\\
&
&
3. Working with the system would be fun.\\
&
&
4. I would like to work with the robot.\\

\midrule

\textbf{Self-Efficacy}~\cite{Venkatesh2003}
&
Assesses caregivers’ confidence in their ability to use care robots effectively under different support conditions.
&
\textit{I could complete a job or a task using the robot...}\\
&
&
1. If there was no one around to tell me what to do as I go.\\
&
&
2. If I could call someone for help if I got stuck.\\
&
&
3. If I had a lot of time to complete the job for which the robot was provided.\\
&
&
4. If I only had the built-in help facility for assistance.\\

\midrule

\textbf{Behavioral Intention to Use the System}~\cite{Venkatesh2003}
&
Assesses caregivers’ willingness to use care robots in their future clinical practice.
&
1. I would intend to use the robot in the next 12 months.\\
&
&
2. I predict I would use the robot in the next 24 months.\\
&
&
3. I would plan to use the robot in the next 24 months.\\

\midrule

\textbf{Ethical Judgments (CAN model)}~\cite{ReinaresLara2018}
&
Assesses ethical evaluations of care robots, including fairness, morality, and cultural acceptability.
&
\textit{How would you rate the robot’s behavior in the video shown?}\\
&
&
1. Unethical/Ethical; 2. Unfair/Fair; 3. Not morally right/Morally right;\\
&
&
4. Not acceptable to my family/Acceptable to my family;\\
&
&
5. Culturally unacceptable/Culturally acceptable;\\
&
&
6. Not personally satisfying/Personally satisfying;\\
&
&
7. Violates an unwritten contract/Does not violate an unwritten contract.\\

\midrule

\textbf{Overall Rating (own)}
&
Helps evaluate the acceptance of human--robot collaboration.
&
1. Overall, I would rate the robot positively.\\
&
&
2. In general, I would support the integration of this robot in healthcare settings.\\

\bottomrule

\end{tabularx}
\end{adjustbox}
\end{table*}

\subsubsection{Open-Ended Questions on Ethical Considerations}

The online questionnaire included a set of optional open-ended questions for each robot type presented. These questions explored stakeholders’ perspectives on the ethical implementation of healthcare robots and complemented the quantitative findings by providing richer insights into acceptance and ethical considerations. The following were the open-ended questions:
\begin{itemize}
    \item  What are your main reasons for accepting or rejecting the use of this robot in caregiving tasks? You may refer to specific situations, feelings, or expectations you have.
    \item Do you see any risks or ethical concerns in using this robot in healthcare? Please explain.     
    \item Based on your professional experience, what ethical issues do you think should be considered when developing or implementing this robot? 
\end{itemize}

\subsection{Analysis Strategy}

This subsection describes in detail the data analysis procedures employed in the study.
 
\subsubsection{Quantitative Analysis}

Prior to conducting the statistical analyses, the internal consistency of all multi-item scales was assessed to ensure the reliability of the measurement instruments. Following established psychometric practice, composite scale scores were computed only after acceptable levels of internal consistency had been confirmed. Specifically, participants’ responses were aggregated across the items within each scale to obtain scale-level scores. 

Descriptive statistics (means and standard deviations) were then computed for each construct and robot type to provide an initial overview of participants’ evaluations and to inspect the distributional properties of the data. These descriptive results served as the basis for subsequent inferential analyses addressing the study’s research questions. To address \textbf{RQ1} and examine differences in participants’ evaluations across robot types, repeated-measures analyses of variance (RM-ANOVA) were conducted, with robot type specified as a within-subject factor and language as a between-subject factor. Separate models were estimated for each dependent variable, including performance expectancy, attitude toward using the robot, behavioral intention to use, ethical judgment, and overall acceptance. 

Interaction effects between robot type and language were tested to assess whether evaluations of different robot types varied across cultural contexts. When omnibus effects were significant, post hoc pairwise comparisons were conducted using Bonferroni-adjusted tests to control for multiple comparisons. Effect sizes (η² or partial η²) were reported alongside \textit{p}-values.

\subsubsection{Directed Qualitative Content Analysis} 
\label{sec:directed-content-analysis}

\begin{figure}
    \centering
    \includegraphics[width=0.48\textwidth]{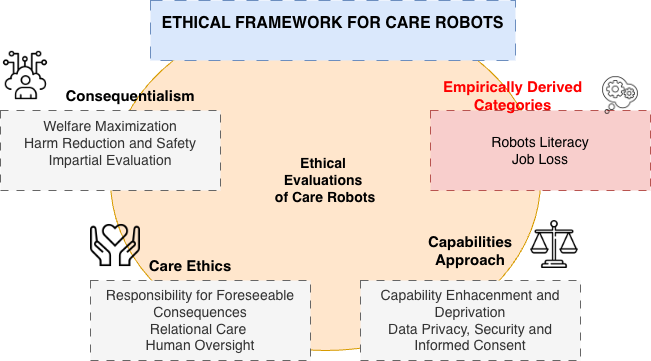}
    \caption{Ethical framework for the evaluation of care robots. The framework combines concepts derived from Consequentialism, Care Ethics, and the Capabilities Approach with empirically derived concerns emerging from participants’ responses to open-ended questions. The circular structure reflects the interrelated and non-hierarchical nature of these perspectives in informing ethical evaluations of care robots.}
    \label{fig:ethical_framework}
\end{figure}

The analysis followed a directed qualitative content analysis approach \citep{hsieh2005three}. A literature-informed ethical framework, drawing on the work of \citep{van2020designing}, was developed to establish deductive analytical categories while allowing for the inductive refinement of themes emerging from the data. To capture the diversity of ethical concerns expressed by participants, the framework integrated three complementary ethical perspectives: care ethics, the capability approach, and consequentialism (see ethical framework in Section~\ref{sec:ethical-perspectives}). This combination enabled the analysis to remain grounded in established ethical theory while remaining sensitive to themes that emerged from the empirical data (see categories in \autoref{fig:ethical_framework}).

The coding process was conducted collaboratively by the first author and a research assistant using QCAmap software. In the first stage, both researchers independently coded 20 percent of the data and subsequently compared and discussed their coding decisions to establish a shared understanding of the coding scheme and refine it where necessary. In the second stage, the first author coded the remaining data and conducted the content analysis. All coding decisions, category assignments, and final interpretations remained the responsibility of the researchers. Through an iterative process of discussion and interpretation, the first and last authors reviewed, refined, named, and defined the themes. Any questions, ambiguities, or disagreements were discussed and resolved through joint meetings until consensus was reached.

\begin{figure}[t]
\centering
\resizebox{0.92\columnwidth}{!}{%
\begin{tikzpicture}[
    font=\sffamily\scriptsize,
    box/.style={
        draw=gray,
        fill=gray!15,
        rectangle,
        align=left,
        minimum width=8.2cm,
        text width=7.8cm,
        minimum height=0.8cm,
        inner xsep=6pt,
        inner ysep=4pt
    },
    arrow/.style={->, thick}
]

\node[box] (s1) {1. Research question, theoretical background};

\node[box, below=0.35cm of s1] (s2)
{2. Definition of the category system from theory (Ethics of Care, Capability Approach, Consequentialism)};

\node[box, below=0.35cm of s2] (s3)
{3. Definition of the coding guideline, containing, for all categories: definitions, anchor examples and coding rules};

\node[box, below=0.35cm of s3] (s4)
{4. Material run-through, preliminary codings. Complementary anchor examples, coding rules};

\node[box, below=0.35cm of s4] (s5)
{5. Revision of the categories and coding guideline\\
after 20\% of the material, comparison coding between researchers};

\node[box, below=0.35cm of s5] (s6)
{6. Final coding and review of the material};

\node[box, below=0.35cm of s6] (s7)
{7. Analysis, category frequencies and\\
contingencies interpretation};

\draw[arrow] (s1.south) -- (s2.north);
\draw[arrow] (s2.south) -- (s3.north);
\draw[arrow] (s3.south) -- (s4.north);
\draw[arrow] (s4.south) -- (s5.north);
\draw[arrow] (s5.south) -- (s6.north);
\draw[arrow] (s6.south) -- (s7.north);

\draw[arrow]
    (s7.west) -- ++(-0.9,0)
    |- (s1.west);

\draw[arrow]
    (s5.east) -- ++(0.7,0)
    |- (s2.east);

\end{tikzpicture}%
}
\caption{Steps involved in the qualitative content analysis.}
\label{fig:qualitative-content-analysis}
\end{figure}
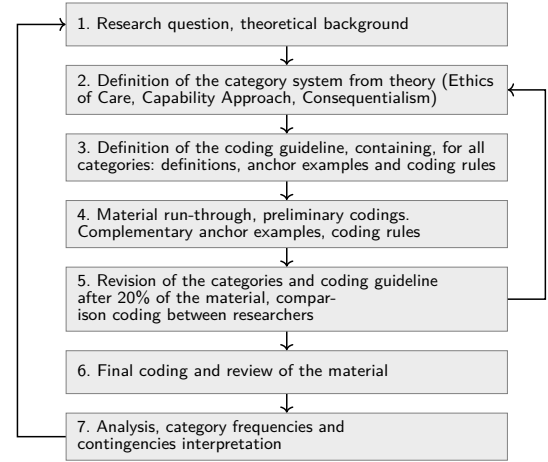

Open-ended responses were originally provided in English and Spanish. To standardize the dataset for qualitative analysis, all Spanish-language responses were translated into English using a single machine-translation system via an API-based scripted workflow in RStudio. Original-language responses were preserved in a separate column for auditability. To enhance translation trustworthiness, translated entries were reviewed by a bilingual researcher.  \autoref{fig:qualitative-content-analysis} illustrates the qualitative analysis process that was carried out. 

\section{Results}

In this section, we present the key findings derived from both the quantitative and qualitative components of the study. Quantitative analyses were conducted using \textit{RStudio} (version 2026.04.0) and \textit{JASP} (version 0.19.3). Qualitative data were analyzed using the web-based application \textit{QCAmap}, in combination with \textit{RStudio}. 

\subsection{Descriptive Statistics }

\autoref{tab:desc} presents the descriptive statistics (means and standard deviations) for all study constructs by robot type and language group. All scales demonstrated acceptable to very good internal consistency across robot types and language versions (Cronbach’s α ranging from .80 to .96). Across constructs and robot types, mean scores were generally above the midpoint of the scale, meaning that, in general, robots were positively rated. Mean values ranged from 3.55 to 6.59, with the lowest scores observed for behavioral intention and the highest scores observed for ethical judgment and overall ratings. Descriptive statistics are reported for participants with complete data and correspond to the sample included in the repeated-measures analyses.

\begin{table*}
\centering
\caption{Descriptive statistics (Means and Standard Deviations) by robot type and language.}
\label{tab:desc}
\footnotesize
\setlength{\tabcolsep}{4pt}
\renewcommand{\arraystretch}{1.1}
\begin{tabular}{l l r r r r r r}
\toprule
Construct & Robot Type 
& EN \textit{M} & EN \textit{SD} & EN $\alpha$
& SP \textit{M} & SP \textit{SD} & SP $\alpha$ \\
\midrule

\multirow{4}{*}{Performance Expectancy}
& DS  & 5.038 & 1.558 & .87 & 5.389 & 1.192 & .87 \\
& BED & 4.678 & 1.661 & .90 & 5.152 & 1.481 & .89 \\
& MVS & 5.003 & 1.553 & .87 & 5.445 & 1.212 & .87 \\
& AM  & 4.831 & 1.655 & .89 & 5.277 & 1.368 & .89 \\

\midrule
\multirow{4}{*}{Attitudes Toward Using Robot}
& DS  & 5.411 & 1.339 & .89 & 5.877 & 1.107 & .87 \\
& BED & 5.115 & 1.559 & .91 & 5.803 & 1.218 & .87 \\
& MVS & 5.360 & 1.506 & .90 & 5.993 & 1.084 & .86 \\
& AM  & 5.457 & 1.361 & .89 & 6.045 & 1.025 & .87 \\

\midrule
\multirow{4}{*}{Self-Efficacy}
& DS  & 5.337 & 1.249 & .84 & 5.267 & 1.221 & .80 \\
& BED & 5.133 & 1.364 & .85 & 5.459 & 1.222 & .85 \\
& MVS & 5.288 & 1.356 & .84 & 5.491 & 1.188 & .84 \\
& AM  & 5.023 & 1.320 & .84 & 5.262 & 1.311 & .84 \\

\midrule
\multirow{4}{*}{Behavioral Intention}
& DS  & 4.851 & 1.912 & .96 & 3.667 & 2.015 & .95 \\
& BED & 4.349 & 2.010 & .96 & 3.744 & 2.040 & .95 \\
& MVS & 4.524 & 2.058 & .95 & 3.779 & 1.969 & .92 \\
& AM  & 4.336 & 2.007 & .95 & 3.546 & 2.016 & .93 \\

\midrule
\multirow{4}{*}{Ethical Judgment (CAN)}
& DS  & 6.109 & 1.103 & .95 & 6.210 & 1.002 & .91 \\
& BED & 5.816 & 1.346 & .95 & 6.147 & 0.969 & .86 \\
& MVS & 5.847 & 1.272 & .95 & 6.070 & 1.200 & .91 \\
& AM  & 6.160 & 1.067 & .96 & 6.450 & 0.814 & .88 \\

\midrule
\multirow{4}{*}{Overall Rating}
& DS  & 5.842 & 1.464 & .88 & 6.209 & 1.210 & .89 \\
& BED & 5.461 & 1.698 & .88 & 6.103 & 1.390 & .89 \\
& MVS & 5.734 & 1.467 & .88 & 6.092 & 1.354 & .89 \\
& AM  & 6.069 & 1.270 & .88 & 6.586 & 0.862 & .89 \\

\bottomrule
\end{tabular}

\smallskip
\footnotesize{Note. Values are scale means; EN = English, SP = Spanish; $\alpha$ = Cronbach’s alpha.}
\end{table*}

\subsection{Quantitative Results}

To examine differences in participants’ evaluations across robot types and to assess whether these evaluations varied as a function of language group, a series of RM-ANOVA were conducted. Robot type (Delivery of Supplies, Helping Patients into Bed, Monitoring Vital Signs, and Assisting with Mobility) was specified as a within-subject factor, and country group (English vs. Spanish) as a between-subject factor. Separate models were estimated for each dependent variable. Specifically, this analysis aims to assess acceptance-related perceptions of four categories of care robots. 

\begin{table*}[t]
\centering
\caption{Repeated-measures ANOVA results across dependent variables.}
\label{tab:rm_anova_results}
\footnotesize
\begin{threeparttable}
\setlength{\tabcolsep}{8pt}
\renewcommand{\arraystretch}{1.2}

\begin{tabular}{>{\raggedright\arraybackslash}p{3.8cm}
                >{\raggedright\arraybackslash}p{3.2cm}
                c
                c
                c
                c}
\toprule
\textbf{Variable} & \textbf{Effect} & \textbf{$F$} & \textbf{$df_n, df_d$} & \textbf{$p$} & \textbf{$\eta^2$} \\
\midrule

\textbf{Performance Expectancy}
& Robot Type & $5.714$ & $3, 888$ & $< .001$ & $.007$ \\
& Language & $10.297$ & $1, 296$ & $.001$ & $.021$ \\
& Robot Type $\times$ Language & $0.192$ & $3, 888$ & $.902$ & $2.445 \times 10^{-4}$ \\
\addlinespace

\textbf{Attitudes Toward Use}
& Robot Type & $5.159$ & $3$ & $.002$ & $.006$ \\
& Language & $23.123$ & $1$ & $< .001$ & $.049$ \\
& Robot Type $\times$ Language & $0.671$ & $3$ & $.570$ & $7.339 \times 10^{-4}$ \\
\addlinespace

\textbf{Behavioral Intention}
& Robot Type & $4.009$ & $3$ & $.008$ & $.003$ \\
& Language & $17.306$ & $1$ & $< .001$ & $.041$ \\
& Robot Type $\times$ Language & $3.306^{*}$ & $3$ & $.020^{*}$ & $.003^{*}$ \\
\addlinespace

\textbf{Self-Efficacy}
& Robot Type & $4.339$ & $3, 888$ & $.005$ & $.005$ \\
& Language & $2.065$ & $1, 296$ & $.152$ & $.005$ \\
& Robot Type $\times$ Language & $3.005^{*}$ & $3, 888$ & $.030^{*}$ & $.003^{*}$ \\
\addlinespace

\textbf{Ethical Judgment (CAN)}
& Robot Type & $15.367$ & $3$ & $< .001$ & $.016$ \\
& Language & $4.910$ & $1$ & $.027$ & $.011$ \\
& Robot Type $\times$ Language & $1.465$ & $3$ & $.222$ & $.002$ \\
\bottomrule
\end{tabular}

\begin{tablenotes}[flushleft]
\footnotesize
\item \textit{Note}. RM-ANOVA = repeated-measures analysis of variance. Effect sizes are reported as $\eta^2$. Greenhouse--Geisser corrections were applied where the assumption of sphericity was violated (Performance Expectancy, Self-Efficacy, and Ethical Judgment). $^{*}$ Indicates statistically significant interaction effects between Robot Type and Language.
\end{tablenotes}
\end{threeparttable}
\end{table*}

The RM-ANOVA results (see Table~\ref{tab:rm_anova_results}) indicate that robot type had a consistent and statistically significant effect across all dependent variables, suggesting that participants’ evaluations systematically varied according to the type of care robot \citep{chatzoglou2024factors, niemela2019social}. Language also showed significant main effects for most outcomes, with the exception of self-efficacy, indicating overall differences between language groups in the perceived acceptance and evaluation of care robots. 

Overall, both language groups tended to evaluate the different robot types in relatively similar ways, as reflected by the largely non-significant interaction effects between Robot Type and Language. These findings point to a general cross-cultural consistency in perceptions of healthcare robots, suggesting that participants from different cultural contexts shared comparable evaluations of the various robotic applications. Moreover, although several effects reached statistical significance, their small effect sizes indicate that the practical magnitude of these differences was limited.\looseness=-1

An exception to this pattern emerged for behavioral intention and self-efficacy, where significant interaction effects between robot type and language were observed. These findings suggest that, while attitudes toward care robots were generally positive across groups, participants differed in their willingness to use these technologies and in their perceived ability to do so. Specifically, participants in the English-language group reported higher behavioral intentions to use care robots than participants in the Spanish-language group. Given the composition of the sample, these differences may reflect broader distinctions between the U.S. context and the Latin American contexts included in the study. Taken together, the results suggest that sociocultural context may play a more prominent role in shaping anticipated adoption and perceived competence in using care robots than in influencing broader attitudinal evaluations.

Bonferroni-adjusted post hoc comparisons (see \autoref{tab:posthoc_results}) revealed that, in the performance expectancy dimension, robots helping patients into bed (BED) received less favorable evaluations than robots delivering supplies (DS) and monitoring vital signs (MVS). In the ethical judgment dimension (CAN model), assisting with mobility (AM) robots received more favorable evaluations than BED and MVS robots, while DS robots were also evaluated more positively than BED and MVS robots. Although descriptive trends (see \autoref{tab:desc}) suggested lower evaluations for BED robots in attitudes toward using robots, Bonferroni-adjusted pairwise comparisons did not reveal consistent significant differences across all robot categories.

Overall, these comparisons suggest that healthcare workers differentiated between forms of physical assistance, evaluating robots involved in direct patient handling less favorably than robots performing logistical or monitoring functions. Likewise, this distinction suggests that not all physically assistive robots are perceived equally, and that the acceptability of robotic care may depend on the specific nature of the caregiving task being performed. These findings support prior research suggesting that acceptance of care robots is task-dependent and shaped by the specific caregiving functions performed by the technology \citep{chatzoglou2024factors, niemela2019social, Formosa2021RobotAutonomy, Johnston2022EthicalDesign}.\looseness=-1

\begin{table}
\centering
\caption{Significant Bonferroni-adjusted post hoc comparisons.}
\label{tab:posthoc_results}
\footnotesize
\setlength{\tabcolsep}{4pt}
\renewcommand{\arraystretch}{1.1}
\begin{tabular}{p{2.5cm} p{1.7cm} p{1cm} p{1cm}}

\toprule
\textbf{Outcome} & \textbf{Comparison} & \textbf{MD} & \textbf{$p_{\mathrm{bonf}}$} \\
\midrule

\textbf{Performance Expectancy}
& DS $>$ BED & $0.298$ & $.004$ \\
& BED $<$ MVS & $-0.309$ & $.002$ \\

\addlinespace[3pt]

\textbf{Ethical Judgment (CAN)}
& DS $>$ BED & $0.178$ & $.029$ \\
& DS $>$ MVS & $0.201$ & $.004$ \\
& DS $<$ AM & $-0.145$ & $.041$ \\
& BED $<$ AM & $-0.323$ & $<.001$ \\
& MVS $<$ AM & $-0.346$ & $<.001$ \\

\bottomrule
\end{tabular}

\smallskip
\footnotesize{\textit{Note}. Only statistically significant Bonferroni-adjusted pairwise comparisons are presented. Mean differences retain the original comparison direction reported in the post hoc analyses. Symbols indicate the direction of comparatively more favorable evaluations between robot categories. DS = Delivering Supplies; BED = Helping Patients into Bed; MVS = Monitoring Vital Signs; AM = Assisting with Mobility.}
\end{table}

\subsection{Qualitative Results}

In the following section, we present the most salient qualitative findings of the study. The analysis is organized into three parts. First, we examine participants’ ethical perceptions of care robots at a general level. Second, we explore how ethical evaluations varied according to the specific characteristics, functions, and tasks performed by the different robot categories. Finally, we analyze cross-cultural similarities and differences in participants’ ethical assessments, highlighting how sociocultural contexts shaped the interpretation, prioritization, and negotiation of ethical concerns surrounding care robot implementation.

\subsubsection{Ethical Themes in Participants' Responses}

\autoref{tab:rotated} provides an overview of the ethical considerations identified in participants’ responses regarding the acceptance and potential rejection of care robot technologies. To analyze these responses, we draw on the categories derived from the theoretical framework described in \cref{sec:directed-content-analysis}. \autoref{tab:rotated} therefore identifies both the theoretical origins of each category and those that emerged inductively during the coding process.

\begin{sidewaystable*}
\centering
\footnotesize
\setlength{\tabcolsep}{3pt}
\caption{Overview of ethical categories, analytical interpretations, and ambivalent evaluations across participants' responses.}\label{tab:rotated}
\begin{tabular}{L{2.3cm} L{2.8cm} L{5.6cm} L{4.5cm} L{4.7cm}}
\toprule
\textbf{Category} & \textbf{Definition} & \textbf{Analytical interpretation} & \textbf{Positive/Neutral quote} & \textbf{Critical/Neutral quote} \\
\midrule
\textbf{Welfare maximization} (\hyperlink{consequentialism}{Consequentialism})
& Ethical arguments emphasizing that care robots are perceived as desirable when they enhance efficiency, improve well-being, and reduce workload, and as undesirable when they fail to contribute meaningfully to these aspects. 
& This category was among the most frequently invoked across both cultural contexts. Participants commonly framed robots as tools for improving efficiency and reducing workload by streamlining routine tasks. However, they also questioned whether these anticipated benefits would be realized in practice or whether robot adoption might instead introduce new forms of workplace disruption. 
& ``It would free up a person to work on something else''. \textit{(Robots Delivering Supplies)}\par
``I consider that this type of robot is extremely necessary and beneficial in the health area. It could help a lot of patients. In addition, it would also benefit caregivers who usually rely on their own physical strength for these tasks.'' \textit{Robots Assisting with Mobility).} 
& ``Although the robot helps me to make the movement and to load the patient, it still has to be watched at the time of performing the action, so I can not be so productive because I have to be watching that the movement is done correctly and that it does it a little slow, so I could lose time if it is done in a place with many patients.'' \textit{(Robot Helping Patients into Bed)}\par
``I feel that taking vital signs myself is faster than using the robot. It would also only be usable with patients who can actively participate in the process.'' \textit{(Robots Taking Vital Signs)}.\\

\addlinespace
\textbf{Harm reduction and safety} (\hyperlink{consequentialism}{Consequentialism})
& Ethical considerations emphasizing that care robots are morally justified insofar as they reduce risk, prevent harm, or minimize negative outcomes. Conversely, they are unjustified if they compromise individuals' well-being by causing accidents or harm. 
& Safety, alongside welfare maximization, emerged as one of the most salient moral concerns. Participants viewed care robots positively when they were perceived as reducing errors, physical strain, burnout, or harm to patients. Conversely, robots were also seen as potential sources of accidents, malfunctions, or loss of control, particularly during physically intimate care tasks. Participants further stressed that safety depends not only on technical performance but also on caregivers’ responsible use of these systems.
& ``I believe it would facilitate the transfer of overweight patients and prevent accidents/injuries to nurses, orderlies and physiotherapists.'' \textit{(Robots Assisting with Mobility)}. \par 
``I think it is a wonderful advancement in technology. It is incredibly hard to support the weight of a person and assist in mobility and it is also riskier for injuries to both client and caregiver without a device like this. \textit{(Robots Assisting with Mobility)}.
& ``I would only be afraid of it hurting, dropping someone, or malfunctioning.'' \textit{(Robots Helping Patients into Bed)}\par 
``Robots need thorough testing to prevent malfunctions that might delay or misdeliver important items like medications or lab samples .'' \textit{(Robots Delivering Supplies)} \\

\addlinespace

\textbf{Impartial evaluation} (\hyperlink{consequentialism}{Consequentialism})
& Ethical arguments emphasizing fairness, equal consideration, and balanced evaluation across stakeholders.
& This category captured responses in which participants evaluated care robots in terms of their consequences for multiple stakeholders. Participants expressed concerns about distributive justice, questioning whether robotic technologies would reinforce or reduce existing inequalities. In particular, they highlighted unequal access to robot technologies. Participants also raised concerns about algorithmic bias, emphasizing that AI systems trained on biased data may produce discriminatory outcomes.
& ``I don't see how the robot can pose any ethical problems as long as it treats each person equally.'' \textit{(Robots Delivering Supplies)}\par
``I think as long as the robot remains neutral and it doesn't assess the patient based on anything that would be considered as being prejudiced or biased, there would be no problems with ethics'' \textit{(Robots Helping Patients into Bed)}.
& ``Equity and accessibility should be addressed. The technology should not only be available to well-funded hospitals but also accessible to smaller facilities and diverse patient populations.'' \textit{(Robots Helping Patients into Bed)}. \par
``Fairness. Developers must ensure that the robot does not reinforce biases or discriminate against any group of patients.'' \textit{(Robots Delivering Supplies)}.\\
\bottomrule
\end{tabular}
\noindent\textit{Note.} Categories identified inductively from the data are indicated with $^{*}$.
\end{sidewaystable*}

\begin{sidewaystable*}
\centering
\footnotesize
\setlength{\tabcolsep}{3pt}
\begin{tabular}{L{2.3cm} L{2.8cm} L{5.6cm} L{4.5cm} L{4.7cm}}
\toprule
\textbf{Category} & \textbf{Definition} & \textbf{Analytical interpretation} & \textbf{Positive/Neutral quote} & \textbf{Critical/Neutral quote} \\
\midrule
\textbf{Responsibility for foreseeable consequences} (\hyperlink{care-ethics}{Care Ethics})
& Ethical arguments emphasizing anticipation of future risks, unintended effects, and responsibility for foreseeable outcomes.
& Participants used this category to express a future-oriented ethical stance centered on preventing foreseeable risks associated with care robots. They emphasized that hospitals, designers, and healthcare institutions should establish clear protocols and response procedures before implementation to mitigate potential harms. Participants also raised concerns about accountability in complex socio-technical systems, particularly when responsibility for errors or adverse outcomes may become diffuse or unclear. 
& ``Hospitals need systems in place to track deliveries accurately and assign responsibility if errors occur, such as delivering the wrong medication.'' \textit{(Robots Delivering Supplies)} \par
``Accountability: Is the manufacturer, the hospital, the managing therapist, or the robot's programmer legally and morally liable if a system malfunctions and a child is hurt?  Prior to implementation, it is necessary to establish clear lines of accountability.'' \textit{(Robots Assisting with Mobility)}
& ``Accountability matters if a robot malfunctions during a transfer who is responsible? The nurse?The hospital questionmark? the manufacturer questionmark? ethical use demands crystal clear accountability and fail-safes because one error could mean a serious injury'' \textit{(Robots Helping Patients in Bed)} \par
``What if the robot supplies the wrong drugs to the wrong room. Who takes the blame?'' \textit{(Robots Delivering Supplies)} \\

\addlinespace

\textbf{Relational care and human oversight}
(\hyperlink{care-ethics}{Care Ethics}) 
& Ethical arguments emphasizing human relationships, emotional connection, interpersonal presence, and appropriate human oversight.
& Participants frequently emphasized that robots should support rather than replace human caregivers. Responses reflected a tension between the efficiency of robotic systems and the ethical importance of empathy, emotional warmth, and human presence. While some viewed robots as valuable complements to care, they stressed that meaningful human oversight must remain central and that responsibility for decisions affecting patients' health should always rest with caregivers. 
& ``Human oversight should always be maintained, ensuring staff can intervene if needed'' \textit{Robots Helping Patients into Bed}.\par
``The robot should be designed to support patient dignity, motivation, and emotional well-being without replacing essential human interaction.'' \textit{Robots Helping with Mobility}
& ``Over-reliance on automated monitoring may reduce direct human interaction, which is important for patient comfort, trust, and emotional support.'' \textit{Robots Monitoring Vital Signs} \par
``Patients may feel a loss of personal dignity or autonomy if they are moved by a machine instead of a human caregiver.'' \textit{(Robots Helping Patients into Bed)}
 \\

\addlinespace

\textbf{$^{*}$Robots literacy} 
& Ethical considerations emphasizing education, training, and understanding of robotic systems.
& Participants emphasized that the successful implementation of care robots depends on providing patients, caregivers, and healthcare stakeholders with clear guidance on how the robot operates and what it is designed to do. Such explanations were considered essential for reducing uncertainty, facilitating appropriate interaction, and establishing realistic expectations. Conversely, insufficient understanding of robotic systems was associated with confusion, misuse, and unrealistic expectations about the robot’s capabilities.
& ``As long as a person is trained, there should not be risks''. \textit{Robots Assisting with Mobility} 
``Just making sure the patient has a good understanding of how the robot will operate in assisting staff with their care.'' \textit{Robots Helping Patients into Bed}\par
& ``I can see where some patients may be fearful of the robot-assisted med cart moving. Educate the patients that this is what is going to be happening.'' \textit{Robots Delivering Supplies}\par
``Risk of lack of staff training.'' \textit{Robots Assisting with Mobility}\\

\addlinespace

\textbf{Capability enhancement and deprivation} (\hyperlink{capability-approach}{Capability Approach})
& Ethical arguments addressing the impact of robots on autonomy, agency, independence, dependence, and opportunities for meaningful participation.
& This category captured the ambivalence surrounding robots as both capability-enhancing and capability-restricting technologies. Some participants viewed robots as tools that could expand patient autonomy, mobility, independence, and participation in healthcare. Others, however, expressed concern that excessive reliance on robotic systems could reduce healthcare workers’ active involvement, weaken professional expertise, or limit human decision-making. 
& ``It seems to me that any tool that is capable of providing a patient with a greater degree of autonomy is a tool that should not be overlooked and should be considered when evaluating the quality of medical care and quality of life of the patient.'' \textit{(Robots Assisting with Mobility)}
``I am in favor of its use to support mobility therapies, help people to walk again and give them a more dignified life.'' \textit{(Robots Assisting with Mobility)} \par
& ``None, they will only make trainees and nurses "lazier", since in the future they may stop teaching how to take vital signs.'' \textit{(Robots Taking Vital Signs)}\par
``Over-reliance on automation: Employees may grow unduly reliant on the robot and fail to notice critical clinical cues that call for human judgment.'' \textit{(Robots Taking Vital Signs)}\\
\bottomrule
\end{tabular}
\noindent\textit{Note.} Categories identified inductively from the data are indicated with $^{*}$.
\end{sidewaystable*}

\begin{sidewaystable*}
\centering
\footnotesize
\setlength{\tabcolsep}{3pt}
\begin{tabular}{L{2.3cm} L{2.8cm} L{5.6cm} L{4.5cm} L{4.7cm}}
\toprule
\textbf{Category} & \textbf{Definition} & \textbf{Analytical interpretation} & \textbf{Positive/Neutral quote} & \textbf{Critical/Neutral quote} \\
\midrule
\textbf{$^{*}$Data privacy, security, and informed consent} (\hyperlink{capability-approach}{Capability Approach})
& Ethical considerations related to sensitive information, data protection, cybersecurity, and consent to interact with robots.
& Participants associated robotic implementation with risks related to data collection, storage, surveillance, and institutional trust. While some viewed the storage of personal data as enhancing safety and security, others perceived it as increasing surveillance and reducing privacy. Participants emphasized that acceptable implementation depends on informed consent and individuals’ freedom to decide whether to use robotic systems. Particular concerns were raised about obtaining informed consent from psychiatric patients and individuals with mental illnesses.
&``It can help to have a more accurate record of patient data without having to annotate, and reduces the risk of human error.'' \textit{(Robots Monitoring Vital Signs)}\par
``None at all. I fully recommend these robots. Of course, we would need to obtain the patient's consent or the family member of the patient if the patient has cognitive impairments such as dementia/Alzheimer.'' \textit{(Robots Helping Patients in Bed)}
& ``Data protection and confidentiality are major concerns. These robots often rely on sensors, cameras, or AI systems that record sensitive health information, so developers must ensure strict data encryption and compliance with healthcare privacy regulations.'' \textit{Robots Helping Patients into Bed}\par
``Data privacy and security is also important, as these robots often operate through centralized systems that track deliveries and item details.'' \textit{(Robots Delivering Supplies)}\\

\addlinespace

\textbf{Contextualization}
(\hyperlink{capability-approach}{Capability Approach})
& Ethical arguments emphasizing adaptation to individual needs, abilities, circumstances, and care contexts.
& Participants rejected one-size-fits-all approaches to robotic implementation, emphasizing that the appropriateness of robots depends on patients, tasks, and care contexts. They expressed particular concern about psychiatric and pediatric settings, where patients may not fully understand or feel comfortable interacting with robotic systems, increasing the risk of distress. Participants also stressed that robots should respect patients’ wishes and preferences and be adapted to the architectural and spatial conditions of hospitals.
&``It is only a robot that assists in mobilization and rehabilitation, so there are not really many ethical issues to comment on, only that the robot is well adapted to each person's mobility needs.'' \textit{(Robots Assisting with Mobility)}\par
``When implementing this in hospitals, they should test different populations within a hospital. That way they could see what type of population receives this type of robot more appropriately and, on the other hand, be able to adapt different types of robots for different populations.'' \textit{(Robots Monitoring Vital Signs)}
&``Care should be adapted to each person, which may be difficult for standardized technology.'' \textit{(Robots Delivering Supplies)} \par
 ``I believe that in specific cases, such as pediatric, geriatric or psychiatric patients, it could become a problem, especially for this public to understand the way in which they are used.'' \textit{(Robots Monitoring Vital Signs)} \\

\addlinespace

\textbf{$^{*}$Job loss} 
& Ethical concerns related to replacement of human labor and socio-economic consequences of automation.
& This category was among the least frequently mentioned, and all responses expressed negative views. Participants raised concerns about job displacement and the impact of robotic technologies on the healthcare workforce. Consistent with the category of relational care and human oversight, many emphasized that robots should support rather than replace caregivers, warning that economic and efficiency-driven incentives could ultimately favor robotic over human care.
& None 
& ``As a matter of ethics, people losing their jobs as they are replaced by robots or AIs.'' \textit{(Robots Delivering Supplies)}\par
``This model can do many things, and I would be a bit concerned about job losses.'' \textit{(Robots Monitoring Vital Signs)}\\
\bottomrule
\end{tabular}
\noindent\textit{Note.} Categories identified inductively from the data are indicated with $^{*}$.
\end{sidewaystable*}

Importantly, the interpretation of these categories does not rely on a simple distinction between positive and negative evaluations. Participants frequently invoked the same ethical considerations to both support and criticize the use of care robots, highlighting the complex and often ambivalent nature of ethical reasoning surrounding these technologies. Overall, the theoretically derived categories captured a substantial proportion of participants’ responses, suggesting that the proposed framework provides a useful lens for understanding stakeholders’ ethical evaluations of care robots. In the following sections, we elaborate on how these ethical considerations emerged across the different analytical categories.

\subsubsection{Cross-Cultural Comparison of Ethical Evaluations}

Based on the qualitative findings, some socio-cultural differences emerged in participants’ ethical perceptions regarding the use of care robots. However, these differences were relatively limited, suggesting that participants across cultural contexts generally identified similar ethical benefits and risks associated with robotic care technologies. Nevertheless, some interpretative differences became evident across specific thematic categories. For instance, among participants from Mexico and Chile, the concept of impartial evaluation was predominantly associated with the economic capacity to acquire or access robotic systems, reflecting concerns related to inequality and accessibility in healthcare technologies. In contrast, participants from the US tended to associate impartial evaluation with concerns regarding algorithmic bias, discriminatory outcomes, and biases embedded within datasets used to train robotic systems. 

Similarly, when discussing data privacy, participants from the US consistently referred to compliance with the Health Insurance Portability and Accountability Act of 1996 (HIPAA), emphasizing the importance of adhering to federal regulations designed to protect the privacy, security, and integrity of Protected Health Information. Although participants from Mexico and Chile also raised concerns regarding data privacy and confidentiality, they generally did not connect these concerns to specific regulatory frameworks or legal standards.

Additional contextual differences were observed in the way participants framed ethical concerns surrounding robotic systems. Responses from the US more frequently emphasized issues related to harm reduction, safety, accountability, and responsibility for foreseeable consequences, suggesting a stronger focus on efficiency, governance, and regulation. In contrast, participants from Mexico and Chile more commonly referred to contextual adaptation,  capability enhancement or deprivation, and the relational dimensions of care, emphasizing the importance of understanding the specific social and emotional realities in which robots are implemented.

One particularly salient finding concerned the use of robots in psychiatric and pediatric care settings. Across both cultural contexts, participants frequently expressed skepticism about implementing robots in these environments. Many emphasized that patients in such settings may have difficulty understanding what robots are or how they operate, potentially resulting in fear, emotional distress, mistrust, or even accidental harm.\footnote{This interpretation is further supported by an exploratory part of the study in which participants were asked to identify the healthcare settings in which care robots would be most useful. Across both cultural contexts, psychiatric settings were consistently evaluated less favorably than other medical environments (see \autoref{fig:Coch2} in the supplementary material).} Overall, the findings suggest that ethical perceptions regarding care robots are shaped not only by universal concerns surrounding healthcare technologies, but also by the social, economic, and regulatory contexts in which these technologies are interpreted and evaluated.

\subsubsection{Ethical Evaluations by Robot Type}

Regarding the comparison between different robotic systems, certain ethical categories were more frequently associated with specific types of robots, suggesting that participants did not evaluate robotic technologies as homogeneous systems, but rather interpreted their ethical implications according to the specific forms of care, task, interaction, and vulnerability involved in each context. Overall, participants appeared to differentiate between robotic systems that support bodily assistance and patient autonomy, and those that mediate clinical judgment, monitoring, or interpersonal interaction.

For example, the category "capability enhancement and deprivation" was particularly prominent in discussions concerning robots assisting with mobility and robots monitoring vital signs; however, the connotations associated with this category differed substantially between the two systems. In the case of robots assisting with mobility, participants generally used this category positively, frequently describing the robot as a tool capable of supporting patient autonomy, improving mobility, and facilitating rehabilitation, particularly in situations involving physical dependence or reduced movement capacity. 

In contrast, discussions surrounding robots monitoring vital signs reflected a predominantly negative interpretation of the same category. Participants frequently associated these systems with the potential loss or deterioration of caregivers’ clinical skills, particularly regarding the manual assessment and monitoring of vital signs. Rather than viewing automation as enhancing, participants often perceived these systems as potentially replacing human competencies and reducing direct clinical engagement. These findings suggest that participants distinguished between forms of automation perceived as augmenting human care and those perceived as substituting essential human skills and interactions. Some illustrative quotes related to these findings can be found in \autoref{tab:rotated}.

Notably, robots assisting with mobility and those helping patients into bed were more frequently associated with promoting dignified care. Participants often emphasized that assistance with mobility and bodily support could enhance patients' dignity by reducing situations in which they feel exposed, dependent, or physically vulnerable in the presence of others. Interestingly, these findings suggest that participants did not necessarily associate robotic intervention with dehumanization; rather, in certain physically sensitive situations, robotic assistance was perceived as potentially less invasive and more dignity-preserving than human intervention alone.

At the same time, robots assisting patients into bed were also strongly associated with the category "harm reduction and safety". Participants frequently expressed concerns regarding the possibility of technical malfunction, mechanical errors, or physical accidents that could directly harm vulnerable patients during bodily transfers. These findings reveal an important ethical tension in participants’ perceptions: while robotic systems were recognized as capable of preserving dignity and reducing physical dependency, they were simultaneously perceived as introducing new forms of physical risk in highly vulnerable care situations. 

Similarly, robots monitoring vital signs and delivering supplies were more frequently associated with the category of relational care and human oversight. Participants emphasized that activities such as monitoring vital signs are not merely technical procedures but also important opportunities for interpersonal connection, emotional support, and human presence. Although these robots were generally viewed positively, participants consistently stressed that their use should remain subject to human and clinical oversight, expressing greater concern about fully autonomous care systems lacking human validation and accountability. 

Overall, the qualitative findings regarding the robot types indicate that participants’ ethical evaluations of care robots were strongly shaped by the type of human relationship affected by the technology. Robotic systems were generally perceived more positively when they supported patient autonomy, reduced physical vulnerability, or preserved dignity, whereas greater skepticism emerged when robots were perceived as replacing human interaction, reducing relational care, or eroding clinical competencies. 

\section{Discussion}

The goal of this study was to investigate caregivers’ acceptance of four categories of care robots across the United States, Mexico, and Chile; and second, to evaluate a literature-informed ethical framework for analyzing stakeholders’ ethical perceptions of these technologies. Three key findings emerged from our analyses. First, participants generally expressed more favorable attitudes toward robots performing logistical and physically demanding tasks than toward robots operating in contexts requiring close interpersonal interaction. Second, although participants from Mexico and Chile reported positive evaluations across most acceptance dimensions, they expressed lower behavioral intentions to use these technologies, suggesting greater uncertainty regarding their practical implementation in local healthcare systems. Finally, the qualitative findings revealed both shared and context-specific ethical concerns related to the design, deployment, and governance of care robots, highlighting how sociocultural values and robot type shape their acceptance.

For the discussion, we focus on integrating the quantitative and qualitative findings to provide a comprehensive understanding of participants’ perceptions, evaluations, and ethical reflections regarding care robots across different sociocultural contexts, as well as to discuss some of the most latent ethical.

\subsection{Task-Dependent Acceptance: Relational and Contextual Dimensions}

Some of the qualitative data provided complementary insights that helped explain participants' quantitative evaluations of care robots. Consistent with previous research \citep{walters2008avoiding,broadbent2009acceptance,degraaf2013exploring,chatzoglou2024factors,niemela2019social}, acceptance varied across the four robot categories, although the overall evaluation patterns remained broadly consistent across cultural contexts. One of the clearest findings was that robots designed to help patients into bed received less favorable evaluations than the other robot types. Qualitative responses suggested that, although participants frequently associated these robots with greater patient autonomy, enhanced dignity, and reduced dependence on caregivers for mobility-related or potentially embarrassing situations, they also perceived them as particularly high-risk technologies. Concerns about accidents, physical harm, loss of control, and technological malfunction appeared to outweigh sometimes their perceived benefits \citep{broadbent2018using}. 

Another prominent quantitative finding was that, within the ethical judgment dimension (CAN model), robots assisting with mobility received more favorable evaluations than robots helping patients into bed or monitoring vital signs. Although all three involve direct patient interaction, mobility-assistance robots are typically perceived as wearable technologies that support, rather than independently perform, caregiving tasks \citep{walters2008avoiding}. This interpretation was reinforced by the qualitative findings, in which mobility-assistance robots were frequently associated with positive ethical outcomes, particularly in relation with increasing autonomy and enhancing quality of life (capability enhancement and deprivation category), whereas robots monitoring vital signs were more often linked to negative ethical evaluations. Together, these findings suggest that ethical evaluations of care robots depend not only on the presence of physical interaction but also on how agency, control, vulnerability, and assistance are embodied within the caregiving relationship. Technologies perceived as supporting patients' autonomy, rather than replacing caregivers, appear to be viewed as more ethically acceptable \citep{Formosa2021RobotAutonomy}. 

Similarly, the quantitative analyses revealed generally limited interaction effects between robot type and language, although behavioral intention varied across language groups for some robot categories \citep{lawrence2025socialnorms,marchesi2021cultural,li2019perceptions}. Participants from Mexico and Chile consistently reported lower behavioral intention scores than those from the United States. One possible explanation for this pattern emerges from the qualitative findings, as participants from Mexico and Chile frequently referred (coded under the category of contextualization) to the high costs of care robots and unequal access to technological resources. These concerns may reflect the challenges faced by countries that rely primarily on imported healthcare technologies, where acquisition, maintenance, and implementation costs constitute substantial barriers to adoption, particularly for healthcare institutions with limited financial and infrastructural resources \citep{schneegans2021race,dutta2024understanding,gereffi2018global,dachs2002inequalities,wall2021ai}.

\subsection{Beyond Instrumental Acceptance: Moral Tensions in Care Robotics}

The qualitative analysis, guided by the theoretical framework proposed in this study, showed that users’ ethical perceptions of care robots are shaped by both the sociocultural context and the robots’ characteristics and functions. Rather than being stable or purely instrumental, ethical evaluations emerged as relational and context-dependent negotiations between competing moral values \citep{vandemeulebroucke2020ethics}. Participants frequently evaluated the same robotic systems from multiple, and sometimes conflicting, ethical perspectives, highlighting the moral complexity of technologies that mediate practices of care \citep{Formosa2021RobotAutonomy}. What some participants perceived as ethically beneficial, others interpreted as ethically problematic or potentially harmful. Next, we discuss some of the most important ethical tensions. 

\subsubsection{Welfare Maximization}

Within the category of welfare maximization, participants acknowledged that robots designed to help patients into bed or monitor vital signs could reduce the physical burden associated with patient handling and continuous supervision, thereby promoting overall well-being. However, they also emphasized that introducing these technologies into clinical workflows may generate new cognitive, technical, and organizational demands. In particular, caregivers would need to learn not only how to operate the robots but also how to coordinate their use within everyday care practices. In this regard, this category was linked to \textit{Robots Literacy}. Consequently, the anticipated gains in efficiency were not perceived as automatic, as the additional workload associated with implementation could partially offset the robots' practical benefits \citep{Greenhalgh2017NASSS,torras2024ethics}. These findings illustrate how values commonly associated with technological innovation, such as efficiency and welfare maximization, become ethically and practically contingent when embedded within the organizational and relational realities of care work \citep{van2016healthcare,van2020designing}.

\subsubsection{Harm Reduction and Safety}

The category of \textit{harm reduction and safety} revealed important ethical tensions regarding the role of automation in caregiving environments. Participants frequently emphasized that care robots could reduce risks and physical strain for healthcare personnel, for example by minimizing medication delivery errors when robots deliver supplies or by preventing back injuries and physical exhaustion when robots help patients into bed. Nevertheless, participants repeatedly expressed concerns regarding the possibility of mechanical failures, malfunctions, or loss of control, which could ultimately generate even greater forms of harm~\citep{Hung2022TechnologicalRisks}.

These findings suggest that participants viewed robotic technologies as shifting, rather than eliminating, risks within the caregiving process. While robots were perceived as reducing certain physical and operational risks, they were also seen as introducing technological vulnerabilities that require continuous human oversight, clear operational protocols, and well-defined accountability structures. Accordingly, concerns about harm reduction and safety were closely intertwined with the categories of \textit{responsibility for foreseeable consequences} and \textit{relational care and human oversight} \citep{van2020designing}.

\subsubsection{Capability Enhancement and Deprivation}

The category \textit{Capability enhancement and deprivation} revealed important tensions in how participants understood autonomy in relation to the use of care robots. On the one hand, some robotic technologies were frequently associated with increased independence, improved mobility, and enhanced support for both patients and caregivers. On the other hand, participants also expressed concerns that excessive reliance on robotic systems could foster new forms of technological dependency that might ultimately undermine autonomy itself \citep{Johnston2022EthicalDesign, Liu2022AutonomyIndependence}. 

Importantly, participants suggested that the increasing delegation of caregiving tasks to robots could contribute not only to the erosion of practical and interpersonal caregiving competencies \citep{ApostolovaLanoix2022}, but also to broader forms of social isolation, loneliness, and the weakening of human interaction within caregiving environments \citep{Turkle2011, Vallor2015, soljacic2024robots}. In this sense, autonomy was not understood as a stable condition, but rather as something relationally constructed and potentially transformed through human–robot interactions. This category was consistently connected to \textit{relational care and human oversight}, suggesting that participants perceived the ethical integration of care robots as dependent not only on technological functionality, but also on the preservation of meaningful human relationships and on users’ ability to critically understand, supervise, and appropriately engage with these systems.
  
\subsection{Implications for Care Robot Design and Implementation}

Based on the findings of the present study, several practical implications can be derived for the design and implementation of care robots.

First, although participants across the examined cultural contexts generally expressed positive attitudes toward care robots, subtle differences in perceptions and concerns may influence their acceptance in diverse sociocultural settings. These differences were particularly evident for robots expected to operate in situations involving close interpersonal interaction and physical proximity, such as those monitoring patients' vital signs or assisting patients into bed. The findings therefore highlight the importance of moving beyond purely technical considerations and adopting a user-centered, culturally sensitive design approach that carefully considers how a robot's appearance, behavior, communication style, and decision-making processes shape users' trust, comfort, and sense of control.

Second, the findings highlight the importance of considering the socioeconomic contexts in which care robots are expected to be deployed. While research on care robotics often emphasizes technical performance and usability, the present study suggests that economic accessibility is also a key determinant of public acceptance. Developers should therefore design care robots with affordability and accessibility in mind, adapting both the technology and its implementation to the realities of different healthcare systems. Importantly, affordability should be regarded not only as a market consideration but also as an ethical design principle, as technologies developed primarily for high-income healthcare systems may exacerbate existing inequalities in access to care.

Third, the findings raise important concerns regarding the use of care robots in mental health and psychiatric settings. Although robotic technologies are increasingly being developed to support individuals with mental health conditions \citep{scoglio2019use,kabacinska2021socially,rabbitt2015integrating},participants feared that robots could increase stress, anxiety, confusion, or emotional distress among vulnerable patients. These findings suggest that deploying care robots in mental health settings should be approached with caution, supported by rigorous risk–benefit assessments and continuous evaluation to ensure patient well-being, particularly for populations that may be disproportionately affected by technological failures or inappropriate interactions \citep{riek2016robotics}.

Finally, the findings of this study suggest that the ethical and practical factors influencing the acceptance of care robots should not be understood as inherently positive or negative. Rather, participants frequently perceived the same robot system as both beneficial and potentially problematic, revealing a fundamental ambivalence in how these technologies are evaluated. This highlights the importance of recognizing and addressing such ambivalence throughout the design process. Instead of focusing exclusively on maximizing functionality or usability, developers should adopt a more holistic and human-centered approach that considers the complex and sometimes conflicting expectations users hold toward care robots.

\section{Limitations}

Several limitations of this study should be acknowledged. First, although statistically significant differences were identified, most quantitative effects were small in magnitude, suggesting that the findings should be interpreted as general tendencies rather than strong predictive relationships. Second, the study focused on participants from the United States, Mexico, and Chile; therefore, the findings cannot be generalized to all sociocultural or healthcare contexts. Additionally, participants evaluated hypothetical care robot scenarios rather than interacting directly with robotic systems in real clinical settings, which may differ from actual experiences of use and implementation. We also only evaluated four types of care robots, however, the interpretation might change with those types. Finally, while the qualitative analysis provided important interpretive insights, the proposed ethical categories remain theoretically situated and interpretive in nature, meaning that alternative analytical perspectives and frameworks could lead to different interpretations of the data.

\section{Conclusion}

Overall, the present analyses demonstrate that caregivers’ acceptance of care robots is shaped by a wide range of interconnected factors, including the specific type of robot, the sociocultural context in which it is introduced, and the characteristics, expectations, and concerns of the different stakeholders involved. The findings further suggest that care robots are not evaluated solely according to their technical functionality or efficiency, but also according to the particular forms of caregiving relationships they mediate. In this sense, participants’ evaluations appeared to be strongly influenced by how different robotic systems configured autonomy, bodily vulnerability, control, intimacy, assistance, and human oversight within caregiving interactions. 

Moreover, the empirical findings revealed the ethical complexity involved in evaluating care robots, as participants frequently articulated contradictory moral evaluations, competing ethical priorities, and tensions between different values. Rather than relying on isolated ethical principles, participants often evaluated these technologies through interconnected ethical values that attempted to account for caregiving practices in a more holistic, relational, and context-sensitive manner.

Ultimately, the findings highlight the importance of developing care robotic technologies that are not only functional and efficient, but also socially meaningful, culturally responsive, and aligned with the values, needs, and lived realities of the communities in which they are intended to operate. More broadly, the study demonstrates that understanding care robot acceptance requires moving beyond instrumental evaluations of technology to a more complex ethical discussion in order to examine how robotic systems transform the social, relational, and ethical dimensions through which care itself is understood and practiced.

\backmatter


\bmhead{Funding}

This work was funded by the Carl Zeiss Foundation with the ReScaLe project and the German Research Foundation (DFG) Emmy Noether Program grant number 468878300.

\bmhead{Data Availability}
The complete dataset and coding materials have been deposited in a public repository and will be made available upon reasonable request.

\section*{Declarations}

\bmhead{Ethics Approval and Consent to Participate}

Before the experiment, participants received all the information and gave informed consent to participate in the study.

\bmhead{Conflict of Interest}

The authors declare that they have no financial or non-financial conflicts of interest.

{
\footnotesize
\bibliography{bibliography}}

\section{Supplementary Material}
\label{appendix 1: Descriptions}

\begin{figure*}[t]
\centering
\includegraphics[width=0.9\textwidth, height=0.35\textheight, keepaspectratio]{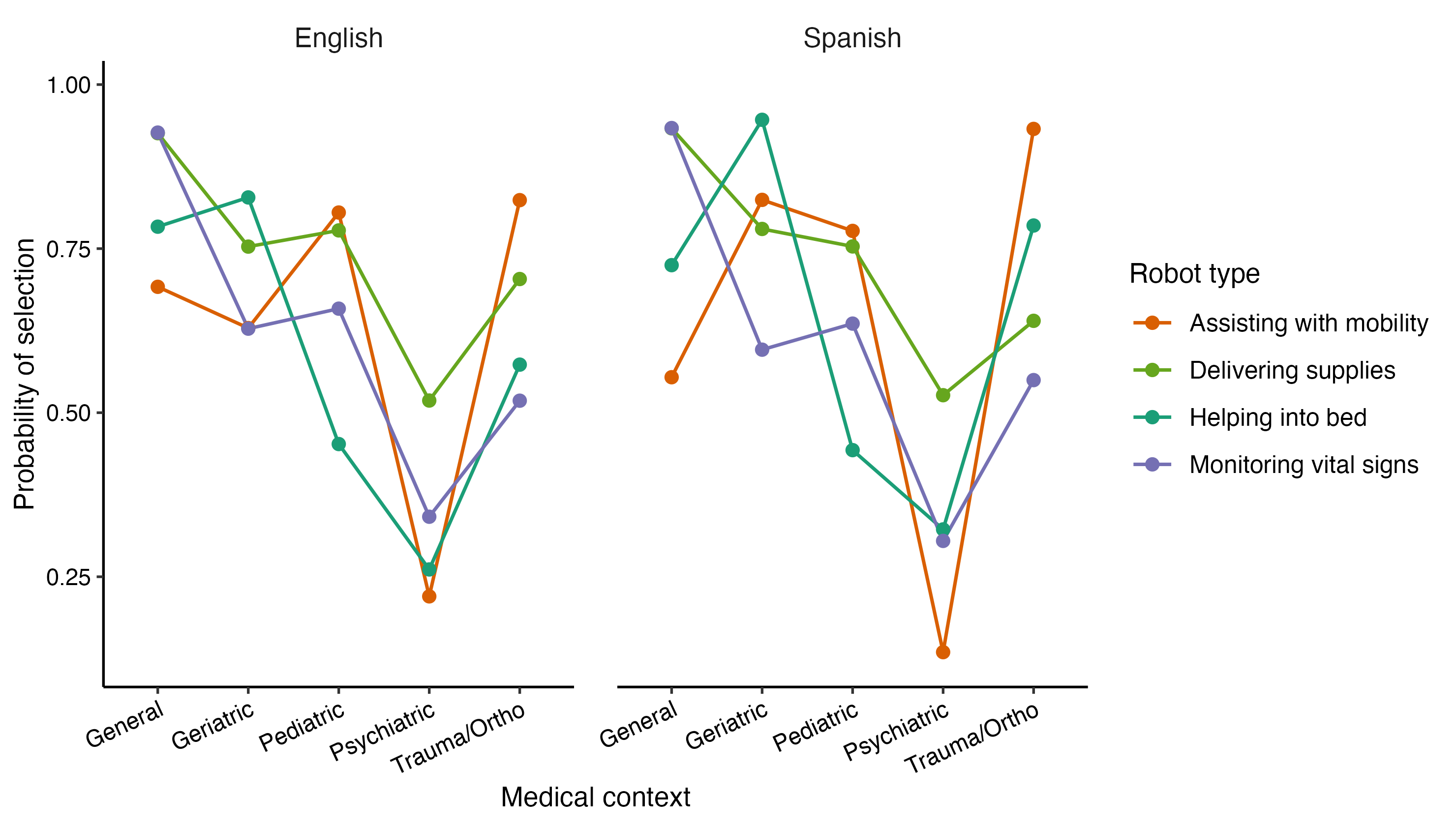}
\caption{Observed proportions of perceived robot suitability across medical contexts by language using Cochran's Q analysis. }
\label{fig:Coch2}
\end{figure*}

As part of the study, participants were also asked to indicate the healthcare settings in which they believed each robot could be appropriately implemented. This question was included to quantitatively examine whether different types of robots were perceived as more suitable for particular healthcare contexts than others (see \autoref{fig:Coch2}). The results revealed a remarkably similar pattern across cultures. In both language groups, robots were consistently perceived as less appropriate for psychiatric settings than for any other healthcare context. This finding complements the qualitative results, in which participants repeatedly expressed concerns that robot deployment in psychiatric environments could increase stress, anxiety, confusion, or emotional distress among vulnerable patients.

\end{document}